\pdfoutput=1

\documentclass[11pt]{article}

\usepackage{EMNLP2023}

\usepackage{times}
\usepackage{latexsym}

\usepackage[T1]{fontenc}

\usepackage[utf8]{inputenc}

\usepackage{microtype}

\usepackage{inconsolata}

\usepackage{algorithm}
\usepackage{algpseudocode}
\usepackage[export]{adjustbox}

\usepackage[mathscr]{euscript}
\usepackage{verbatim}
\usepackage{hyperref}
\usepackage{graphicx}
\usepackage{bm}
\usepackage{amssymb}
\usepackage{amsmath}
\usepackage{makecell}
\usepackage{caption}
\usepackage{subcaption}
\usepackage{stmaryrd}

\newcommand{\pred}{\bm{f}}
\newcommand{\vg}{\bm{G}}
\newcommand{\vx}{\bm{X}}
\newcommand{\vy}{\bm{Y}}
\newcommand{\vu}{\bm{U}}
\newcommand{\vw}{\bm{W}}

\newcommand{\va}{\bm{A}}
\newcommand{\vm}{\bm{M}}
\newcommand{\Var}{\mathbb{V}}

\title{TaCo: Targeted Concept Erasure Prevents Non-Linear Classifiers From Detecting Protected Attributes}

\author{Fanny Jourdan \\
  IRT Saint Exupéry  \\
  Toulouse, France \ \\
  \texttt{fanny.jourdan@irt-saintexupery.com} \\
  \And
  Louis Béthune \\
  Apple \\
  Paris, France \\ 
  \AND
  Agustin Picard \\
  IRT Saint Exupéry \\
  Toulouse, France \\
  \And
  Laurent Risser \\
  IMT, Université Paul-Sabatier \   \\
  Toulouse, France \\ \And
  Nicholas Asher \\
  \ \ IRIT, Université Paul-Sabatier  \\
  Toulouse, France \\
  }

\begin{document}
\maketitle
\begin{abstract}

Ensuring fairness in NLP models is crucial, as they often encode sensitive attributes like gender and ethnicity, leading to biased outcomes. Current concept erasure methods attempt to mitigate this by modifying final latent representations to remove sensitive information without retraining the entire model. However, these methods typically rely on linear classifiers, which leave models vulnerable to non-linear adversaries capable of recovering sensitive information.
 
We introduce Targeted Concept Erasure (TaCo), a novel approach that removes sensitive information from final latent representations, ensuring fairness even against non-linear classifiers.  Our experiments show that TaCo outperforms state-of-the-art methods, achieving greater reductions in the prediction accuracy of sensitive attributes by non-linear classifier while preserving overall task performance.
Code is available on \href{https://github.com/fanny-jourdan/TaCo}{GitHub}.
\end{abstract}

\section{Introduction}
   
Artificial Intelligence models are increasingly deployed in critical domains but often unintentionally encode sensitive attributes such as gender, ethnicity, and age, leading to unfair and unethical outcomes. Ensuring fairness in these models is imperative; and according to information theory, a fair model should be incapable of predicting these sensitive variables, regardless of whether the predictor is linear or non-linear. 

Concept erasure methods have been developed to meet this desideratum. These methods modify the representation in the latent space of NLP models just before classification, 
attempting to remove information about the sensitive variable in this representation, to enable fairer classification without the need to re-train the entire model. 
These methods aim to hide sensitive information in final latent space of models, but generally evaluate their effectiveness using linear classifiers such as logistic regression \citep{INLP, RLACE, LEACE}.  While these approaches reduce the ability of linear models to recover sensitive information, they often fail to eliminate it entirely, leaving it accessible to non-linear models \citep{ravfogel2023loglinear}. This is a significant limitation for many concept erasure methods to achieve true fairness.

To address this limitation, we propose Targeted Concept Erasure (TaCo), a novel concept erasure method that effectively removes sensitive information from final latent representations, ensuring fairness even against non-linear adversaries. Our approach involves three key steps:  \\
\textit{(i) Concept Discovery}: We identify relevant concepts within the latent space that are associated with the sensitive attribute. \\
\textit{(ii) Concept Ranking}: We rank these concepts based on their importance to both the sensitive variable and the target label. This ranking allows us to find the optimal trade-off between fairness and task performance. \\
\textit{(iii) Concept Erasure}: We systematically remove concepts, prioritizing those that are most influential to the sensitive attribute but have minimal impact on the primary task prediction.


We evaluate our method on the \textit{Bios} dataset \citep{de2019bias} using a variety of pre-trained language models, including RoBERTa \citep{liu2019roberta}, DistilBERT \citep{sanh2019distilbert}, DeBERTa \citep{he2020deberta}, and T5 \citep{raffel2020T5}. Our experiments demonstrate that TaCo significantly outperforms state-of-the-art concept erasure methods in preventing non-linear classifiers, such as multilayer perceptrons (MLPs), from predicting sensitive attributes. Specifically, TaCo achieves a greater reduction in the accuracy of predicting sensitive variables across all tested models, without compromising overall (intended) task performance.

In addition to enhancing fairness, our method offers interpretability benefits. By identifying and manipulating specific concepts, we answer the question: \textit{"Which concepts have an influence?"}.
And we further extend our approach by applying concept-based explainability techniques --specifically, COCKATIEL \citep{jourdan2023cockatiel}-- directly on the identified concepts. This allows us to also answer the question: \textit{"Why did this concept have an influence?"}, providing deeper understanding and transparency into the model's decision-making process.

\section{Related work}

In information theory, a common approach to improving the fairness of a model is to eliminate information about sensitive variables to prevent the model from using it  \citep{kilbertus2017avoiding}. This approach to fairness, which we'll call "\textit{sensitive variable information erasure}", has already been used in various forms in the literature

One approach just removes explicit gender indicators (\textit{he}, \textit{she}, \textit{her}, etc.) directly from the dataset~\citep{de2019bias}. However, modern transformer models are able to learn implicit gender indicators~\citep{devlin2018bert,karita2019comparative}, which essentially limits the impact that such a naive approach can have on state-of-the-art models. In fact, a RoBERTa model trained without these gender indicators can still predict with upwards of 90\% accuracy the gender of a person writing a short LinkedIn bio (see Appendix~\ref{apx:genderpredrob}).

A second approach attempts to create word embeddings that do not exhibit gender bias~\citep{bolukbasi2016man, garg2018word, zhao2018gender, zhao2018learning, caliskan2017semantics,bolukbasi2016man,zhao2019gender}. Although interesting in principle, \citet{gonen2019lipstick} shows these methods may camouflage gender biases in word embeddings, but fail to eliminate them entirely. 

A third approach, more in line with our proposed method, consists in intervening on the model's latent space. \cite{liang2020towards} introduce a technique based on decomposing the latent space using a PCA and subtracting from the inputs the projection into a bias subspace that carries the most information about the target concept. 
Another vein focuses on modifying the latent space with considerations coming from the field of information theory \citep{Xu2020A}. Indeed, the $\nu$-information – a generalization of the mutual information – has been used as a means to carefully remove from the model’s latent space only the information related to the target concept. In particular, popular approaches include the removal of information through linear adversarials on the last layer of the model~\citep{INLP,RLACE}, through Assignment-Maximization adversarials~\citep{shao2023erasure}, and through linear models that act on the whole model~\citep{LEACE}. 
However, \citet{ravfogel2023loglinear} has recently shown a limitation of these methods. The limitation comes from the practical use of the $\nu$-information ($\nu$-information is a metric that updates the mutual information, considering the expressivity of functions measuring information in the latent space).
\citet{ravfogel2023loglinear} explains that by protecting information using a linear model, more expressive models can still decode this protected information. Our method avoids this problem, because we are effectively destroying information -- and not simply hiding it from linear models.  Thus, no model can access it. 

In addition, our method, inspired by concept-based explainability techniques like TCAV~\citep{kim2018interpretability}, CRAFT~\citep{fel:etal:2023}, and COCKATIEL~\citep{jourdan2023cockatiel}, provides the added benefit of enabling interpretability of the erased concepts. Unlike most concept erasure methods, which focus solely on removing sensitive information, our approach offers insight into how these concepts influence model predictions, enhancing transparency and interpretability without sacrificing performance.

\begin{figure*}[t]
    \centering
\includegraphics[width=1\linewidth]{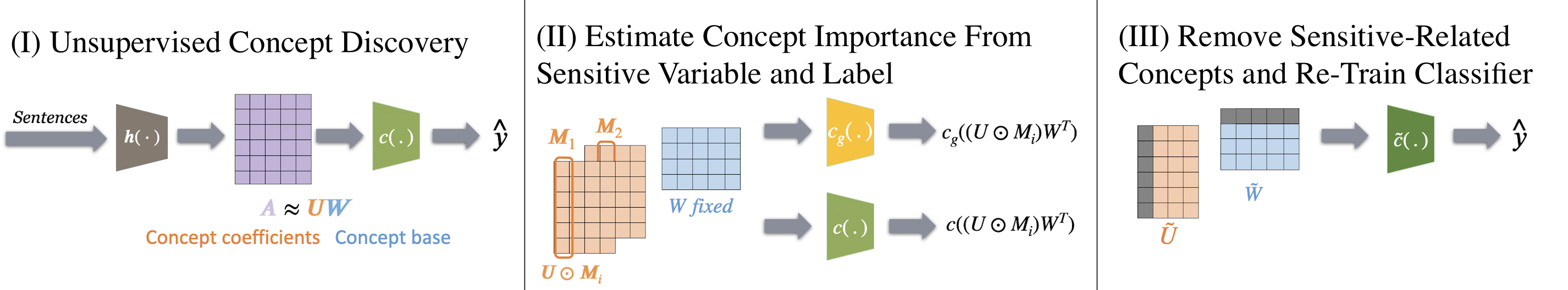}
    \caption{\textbf{Overview of TaCo method.}  A decomposition of the final latent embedding matrix \textbf{(I)} yields concepts, whose importance \textbf{(II)} with respect to the sensitive variable and the label are evaluated with Sobol method. Finally, some concepts are removed, which ``neutralizes'' the sensitive variable information and \textbf{(III)} produces a fairer classifier.}
    \label{fig:diagram}
\end{figure*}

\section{Preliminaries and definitions}

The idea behind our method for concept erasure has 3 components: 
\textit{(i)} identifies label-related concepts within the final latent space of an NLP model (such as the [CLS] embedding for encoder transformers like BERT, the EOS token embedding for encoder-decoder transformers like T5, or the first predicted token for decoder transformers like Llama).   \textit{(ii)} evaluates the importance of each concept for the prediction of the sensitive variable and the label to find the concepts to remove with the best trade-off between removing sensitive information and maintaining model performance; \textit{(iii)} removes  specific concepts to construct a new final latent embedding without sensitive variable information, after which we retrain a classifier to predict label based on this modified embedding. For a visual representation of the method, refer to Figure \ref{fig:diagram}, which provides an explanatory diagram illustrating the different stages of the process.


Our method relies on a mathematical foundation that guarantees removal and ensures the identification and application of suitable, practical tools. This foundation also establishes clear criteria for selecting practical tools and methodologies, enabling the achievement of dependable and consistent results. These are defined in the next section. 


\subsection{Notation}
We work in context of supervised learning assume that a neural network model, denoted as $\pred \colon \mathcal{X} \rightarrow \mathcal{Y}$, has been previously trained to perform a specific classification task. Here, $\bm{x} \in \mathcal{X}$ represents an embedded input text (in our example, a LinkedIn biography), and $y \in \mathcal{Y}$ denotes their corresponding label (in our example, the occupation).  We introduce an additional set, $g \in \mathcal{G}$, representing the sensitive variables (in our example, the gender). This variable is crucial as it is the one we have to control to ensure fairness in our model. With $n$ individuals in the training set, we define $\vx = (\bm{x}_1,...,\bm{x}_n) \in \mathcal{X}^n$, $\vy = (y_1,...,y_n) \in \mathcal{Y}^n$, and  $\vg = (g_1,...,g_n) \in \mathcal{G}^n$.

We represent the model $\pred$ as a composition of two functions. 
\begin{equation}\label{eq:general_pred_model}
\pred(\bm{x}) = c \circ h(\bm{x}) \,,
\end{equation}
where $h(\bm{x}) \in \mathbb{R}^d$. 
$h$ encompasses all layers from the input $\bm{x}$ to a latent space, which is a transformed representation of the input data. The classification layers $c$ classify these transformed data.  $h(\vx)$ is the final latent embedding matrix, $\va = h(\vx) \in \mathbb{R}^{n \times d}$, which is also referred to as the [CLS] embedding in the context of an encoder transformer, as the EOS token embedding for encoder-decoder transformers and the first predicted token for decoder transformers for a classification task. 

\subsection{Causal underpinnings of the method}

A faithful reconstruction of $\pred$ should support a causal relation: $A$ {\em and} $B$ and {\em had} $A$ {\em not been the case,} $B$ {\em would not have been the case either} \citep{lewis:1973,pearl2009causality,jacovi:goldberg:2020}. $A$ describes the causally necessary and ``other things being equal'' sufficient condition for $B$. Faithfulness implies that a particular input type is causally necessary for the prediction at a given data point $\bm{x}$. Such a condition could involve specific variable values or a set of variables and can also be a higher-level logical statement.

A counterfactual theory is a collection of such statements, and for every deep learning model, there is a {\em counterfactual theory} that describes it \citep{jacovi:etal:2021,asher:etal:2022,yin:neubig:2022}. This theory encodes a counterfactual model \citep{lewis:1973} with: a set of worlds or ``cases'' $\mathscr{W}$, a distance metric $\|.\|$ over $\mathscr{W}$, and an interpretation function that assigns truth values to atomic formulas at $w \in \mathscr{W}$ and recursively to complex formulas, including counterfactuals. The cases in a counterfactual model represent elements varied through counterfactual interventions. The distance metric identifies minimal post-intervention elements most similar to the pre-intervention case. We can try various counterfactual models for a given function $\pred$ \citep{asher:etal:2022}.

For example, we can choose:
\textit{(i)} $\mathscr{W} = \mathcal{X}^n$ (model's input entries), \textit{(ii)} $\mathscr{W}$ as the set of possible attention weights in $\pred$ (if it's a transformer model),
\textit{(iii)} $\mathscr{W}$ as the possible parameter settings for a specific intermediate layer of $\pred$ (e.g., attribution matrix as in \citep{fel:etal:2023}),
\textit{(iv)} $\mathscr{W}$ as the possible parameter settings for the final layer of an attention model as suggested in \citep{wiegreffe:pinter:2019}, or
\textit{(v)} sets of factors or dimensions of the final layer embedding matrix from the entire model \citep{jourdan2023cockatiel}.

Our explanatory model follows the ideas of \citep{jourdan2023cockatiel} but differs in details. Notably, our semantic approach allows Boolean combinations, even first-order combinations of factors, to isolate a causally necessary and sufficient {\em explanans}.

\subsubsection{Counterfactual intervention for concept ranking}

In part 2 of our method, we calculate the causal effect on the sensitive variable for each dimension $r$ of $\vu$ and $\vw$ of the $\va=\vu\vw$ decomposition created in part 1 of the method (for more details on this decomposition see section \ref{sec:decomposition}). We conduct a counterfactual intervention that involves removing the dimension or ``concept'' $k \in [\![1;r-1]\!] $. By comparing the behavior of the model with the concept removed and the behavior of the original model, we can assess the causal effect of the concept on the sensitive variable. As explained above, this assumes a counterfactual model where $\mathscr{W}$ corresponds to the possible variations of $\va=h(\vx)$ when reducing the dimensionality of its decomposition $\va=\vu \vw$.

To really able to remove the targeted dimension, the dimensions must be orthogonal of each other. Our mathematical framework guarantees this as far as possible.

\subsection{Information theory guarantees}\label{sec:inftheory}
As part of our counterfactual intervention, we calculate the "importance of a dimension" for the sensitive variable, so that our final integration $\va$ is \textit{as independent as possible} from the sensitive variable $\vg$. According to information theory, if we are unable to predict a variable $\vg$ from the data $\va$, it implies that this variable $\vg$ does not influence the prediction of $y$ by $\va$~\citep{Xu2020A}. This definition is widely used in causal fairness studies \citep{kilbertus2017avoiding}, making it a logical choice for our task.

To evaluate the independence between $\mathbf{a}$ and $\mathbf{g}$, we train a classifier $c_g$ to predict the sensitive variable $\mathbf{g}$ from the representation $\mathbf{a}$ (see Appendix~\ref{apx:modelsdetailsclassif} for architectural details). If $c_g$ cannot achieve high accuracy in predicting $\mathbf{g}$, it suggests that $\mathbf{a}$ lacks information about the sensitive attribute.

While existing methods often employ linear classifiers to estimate this predictability, focusing solely on linear dependencies may overlook non-linear relationships, allowing residual sensitive information to remain accessible to non-linear models~\citep{ravfogel2023loglinear}. To address this limitation, we utilize a non-linear classifier—specifically, a multilayer perceptron (MLP)—to estimate the predictability of $\mathbf{g}$ from $\mathbf{a}$. This approach enables us to detect and mitigate both linear and non-linear dependencies, providing a more robust assessment of the residual sensitive information.

Calculating the importance of each dimension by individually removing them and retraining the classifier can be computationally intensive, especially in high-dimensional spaces. To circumvent this, we train a single non-linear classifier on the original representation $\mathbf{a}$ and apply a sensitivity analysis method (Section~\ref{sec:importance}) to determine each dimension's importance for predicting $\mathbf{g}$. This strategy efficiently identifies the dimensions most influential to the sensitive attribute without the need for exhaustive retraining.

For quantifying the degree of dependence between $\mathbf{a}$ and $\mathbf{g}$, we adopt the $\nu$-information metric. While this metric has been well-established in prior work using linear classifiers~\citep{ravfogel2023loglinear,LEACE}, we extend its application by incorporating non-linear classifiers in its computation. This extension allows us to capture non-linear dependencies that linear models might miss, offering a more comprehensive measure of mutual information.

By using a non-linear classifier to compute $\nu$-information, we provide a stronger guarantee of fairness against adversaries employing complex, non-linear models. In Section~\ref{sec:results-analysis-quantitative}, we show that our method significantly reduces the presence of sensitive information, as evidenced by the decreased predictability of $\mathbf{g}$ even when using sophisticated non-linear classifiers.



\section{Methodology}
Now that we've established constraints on how to implement our method, we define the tools used for each part of the method. \textit{(i)} For the decomposition that uncovers our concepts, we use different decompositions: SVD, PCA, and ICA. \textit{(ii)} For the sensitive variable importance calculation, we train a classifier for this task and then use Sobol importance. \textit{(iii)} To remove concepts, we simply remove the columns of the matrix that correspond to the target concept. 

\subsection{Generating concepts used for occupation classification}\label{sec:decomposition}

Firstly, we examine the latent space before the last layer of the model. We have an final latent embedding $\va \in \mathbb{R}^{n \times d}$ of the transformer model in this space.
The aim here is to employ matrix decomposition techniques to extract both a concept coefficients matrix $\vu$ and a concept base matrix $\vw$ for the latent vectors.

To illustrate this part, we discover concepts without supervision by factorizing the final embedding matrix using different decompositions: singular value decomposition (SVD), principal component analysis (PCA)~\citep{wold1987principal} 
and independent component analysis (ICA)~\citep{comon1994independent}.





For \textbf{SVD}, the matrix $\va$ is decomposed into $\va = U_0 \Sigma_0 V_0^\intercal$, where $U_0 \in \mathbb{R}^{n \times n}$ and $V_0 \in \mathbb{R}^{d \times d}$ are orthonormal matrices, and $\Sigma_0 \in \mathbb{R}^{n \times d}$ is diagonal. This decomposition reveals the main variability in $\va$. By retaining the $r \ll d$ largest singular values in $\Sigma$, we approximate $\va$ as:
\begin{equation}\label{eq:decomp_A}
\va \approx \vu \vw \,,
\end{equation}
where $\vu \in \mathbb{R}^{n \times r}$ contains the leading $r$ columns of $U_0$, and $\vw = \Sigma V^\intercal \in \mathbb{R}^{r \times d}$ combines the singular values and right singular vectors (see \citep{EckartYoung1936}). SVD is used to capture the most significant patterns in the data by reducing dimensionality while preserving maximum variance.

In \textbf{PCA} and \textbf{ICA}, $\va$ is projected onto the directions of maximum variance, resulting in a similar approximation:
\[
\va \approx \vu \vw \,,
\]
where $\vu \in \mathbb{R}^{n \times r}$ contains the top $r$ principal components (the eigenvectors of the covariance matrix of $\va$) for the PCA or contains $r$ independent components for the ICA, and $\vw \in \mathbb{R}^{r \times d}$ contains the corresponding loadings for the PCA or is the mixing matrix for the ICA.

We use PCA to reduce dimensionality by capturing as much variance as possible with the fewest components, making it useful for data compression and noise reduction. And ICA is used to identify underlying factors or sources in the data that are mutually independent, revealing hidden structures not accessible via PCA or SVD.

By expressing $\va$ as a product of $\vu$ and $\vw$ across these methods, we obtain lower-dimensional representations that capture essential characteristics of the data while emphasizing different statistical properties: SVD focuses on capturing overall variability, PCA emphasizes variance and decorrelation, and ICA highlights statistical independence.

Note that the ICA does not give orthogonal dimensions, as we requested in the decomposition prerequisites, but we decide to study it here anyway, in the hope that the statistical independence between its dimensions may bring interesting results. 

For more details on the implementation, we refer the reader to Appendix~\ref{apx:decompo_details}.

\subsection{Estimating concept importance with Sobol Indices}\label{sec:importance}

We have seen in Eq. \eqref{eq:general_pred_model} that $c$ is used to predict $\vy$ based on $\va$, and we discussed in section \ref{sec:inftheory}, that the non-linear classifier $c_{g}$ also uses $\va$ as input but predicts the gender $\vg$. We also recall that $\va$ is approximated by $\vu\vw$ in Eq. \eqref{eq:decomp_A}, and that the columns of $\vu$ represent the so-called \emph{concepts} extracted from $\va$.  

We then evaluate the concepts' importance by applying a feature importance technique on $\vg$ using  $c_{g}$, and on $\vy$ using $c$. More specifically, Sobol indices~\citep{sobol1993sensitivity} are used to estimate the concepts' importance, as in~\citep{jourdan2023cockatiel}. A key novelty of the methodology developed in our paper is to simultaneously measure each concept's importance for the outputs $\vg$ and $\vy$, and not only $\vy$ as proposed in~\citep{jourdan2023cockatiel}.

\begin{figure}[h]
    \centering
\includegraphics[width=1\linewidth]{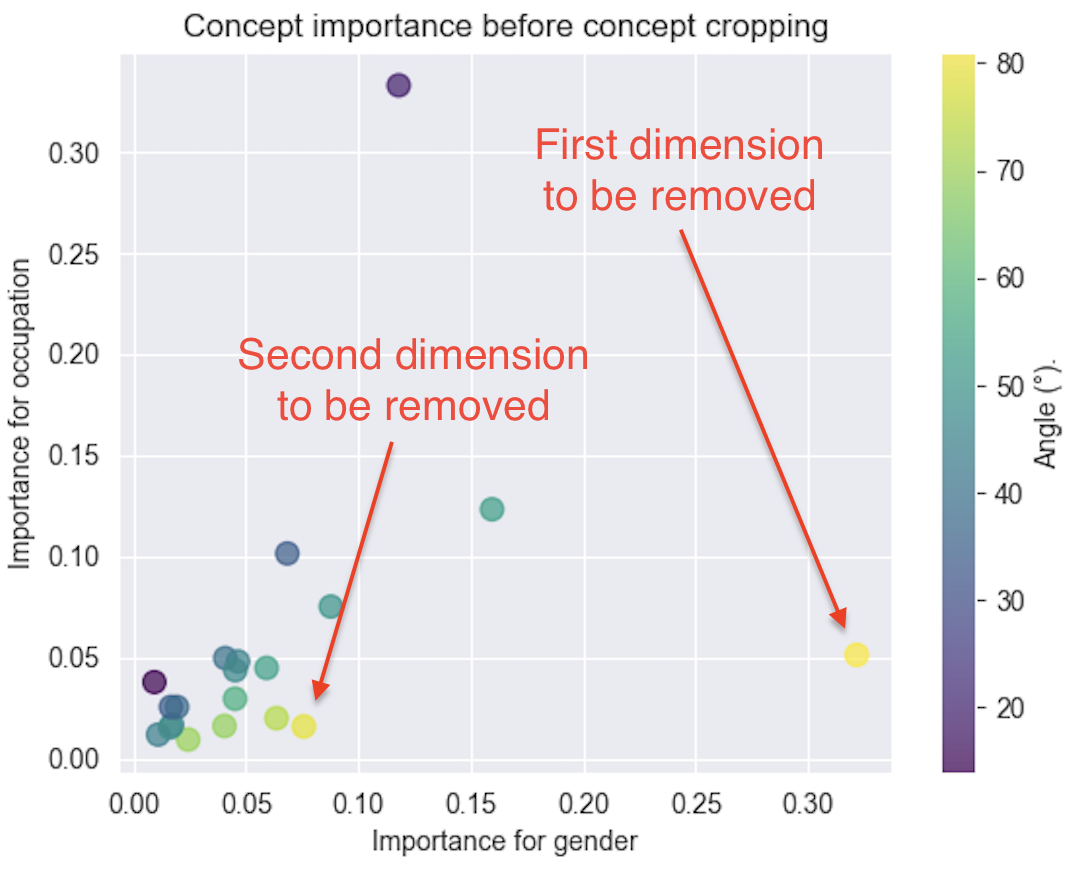}
    \caption{
    \textbf{Co-importance plot for $r=20$ dimensions with respect to \textit{Occupation} and \textit{Gender} labels, computed with the Sobol method on RoBERTa model.} The color is based on  the angle $a=90-\frac{2}{\pi}\arctan{(\frac{y}{x})}$. The angles with extreme values correspond to concepts of high importance for gender but comparatively low importance for occupation. Their removal yields a favorable tradeoff in accuracy/fairness.
    }
    \label{fig:conceptimp}
\end{figure}

To estimate the importance of a concept $\vu_i$ for sensitive variable (resp. for the label), we measure the fluctuations of the model's output $c_{g}(\vu \vw)$ (resp. $c(\vu \vw)$) in response to perturbations of concept coefficient $\vu_i$.

We then propagate this perturbed activation to the model output $\vg = c_{g}(\tilde{\va})$ (resp. $\vy = c(\tilde{\va})$). We can capture the importance that a concept might have as a main effect -- along with its interactions with other concepts -- on the model’s output by calculating the expected variance that would remain if all the concepts except the $i$-th were to be fixed. This yields the general definition of the total Sobol indices.
An important concept will have a large variance in the model's output, while an unused concept will barely change it. 

We explain the total Sobol indices for each concept in detail in Appendix~\ref{apx:sobol}, along with our perturbation strategy.

\subsection{Neutralizing the embedding with importance-based concept removal} 

In this final section, we propose to eliminate $k$ concepts that exhibit significant dependence on sensitive variables from the final embedding matrix. This provides us with an improved embedding which can serve as a base to learn a fairer classifier to predict downstream tasks and, thus, a model that is globally more neutral with regard to sensitive variables.

This concept-removal procedure consists of choosing the $k$ concepts that maximize the ratio between the importance for sensitive variable prediction and for the task. This graphically corresponds to searching for the concepts whose angle with respect to the line of equal importance for sensitive variable and task prediction are the highest (see Fig.~\ref{fig:conceptimp}). This strategy allows us to work on a trade-off between the model's performance and its fairness.

In practice, we sort the concepts $\vu_1$,$\vu_2$,....$\vu_r$ according to this strategy in descending order and delete the first $k$ concepts. This essentially leaves us with a concept base $\tilde{\vu} = (\vu_{k+1},..., \vu_r)$. This same transformation can be applied to $\vw$'s rows, yielding $\tilde{\vw}$.
From the matrix obtained from the matrix product $\tilde{\vu} \tilde{\vw}$, we retrain a classifier $\tilde{c}$ to predict $\vy$ (refer to Appendix~\ref{apx:modelsdetailsclassif} for more details).

It is interesting to note that, because we are causally removing information from the final latent space, it is impossible for any model to recover it from the truncated space -- \textit{i.e.} we are not limited by the family of models that's intervening~\citep{ravfogel2023loglinear}. 

\begin{figure*}[h]
    \centering
    \begin{subfigure}{.5\textwidth}
        \centering
        \includegraphics[width=1\linewidth]{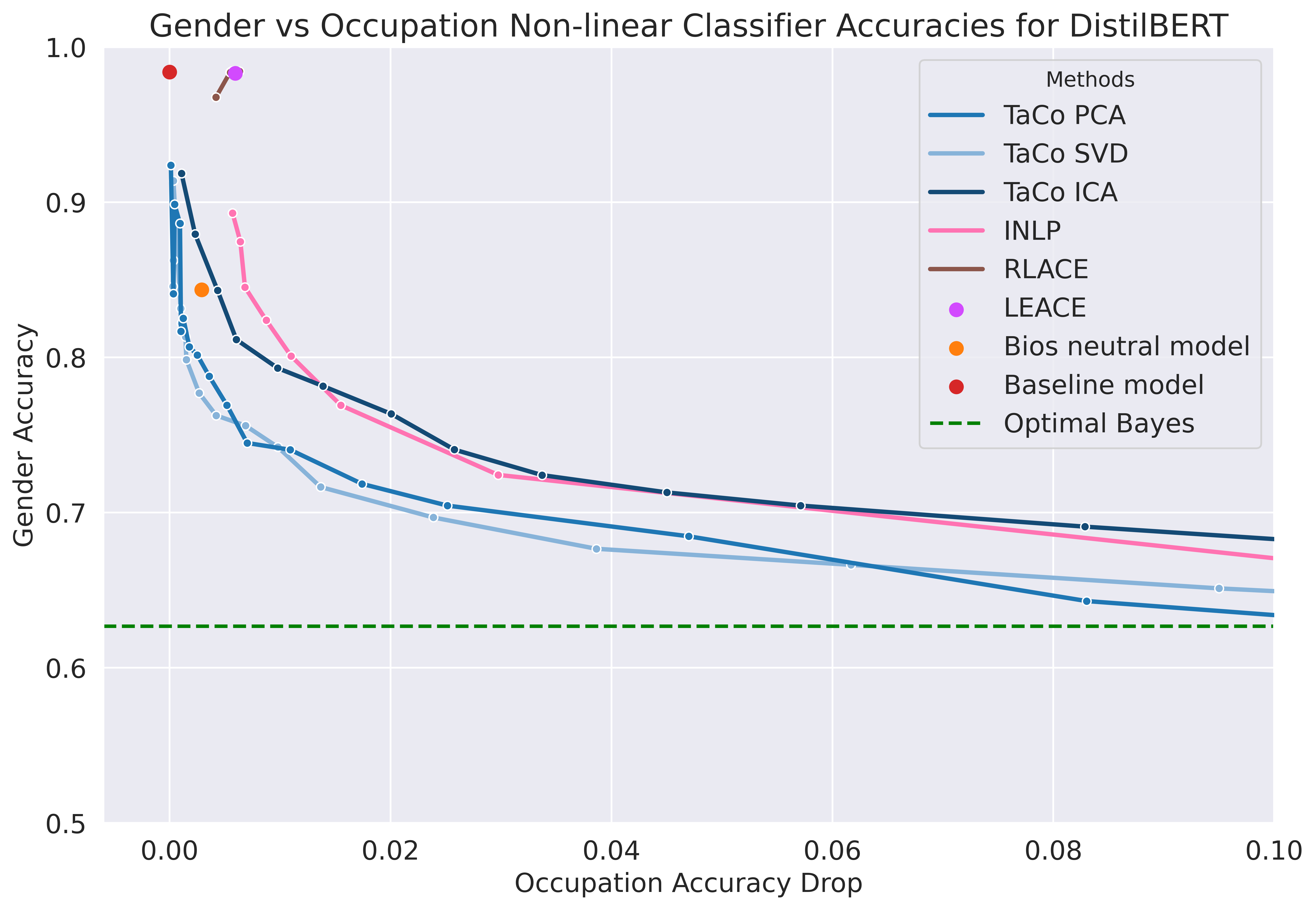}
    \end{subfigure}%
    \hspace*{\fill}
    \begin{subfigure}{.5\textwidth}
        \centering
        \includegraphics[width=1\linewidth]{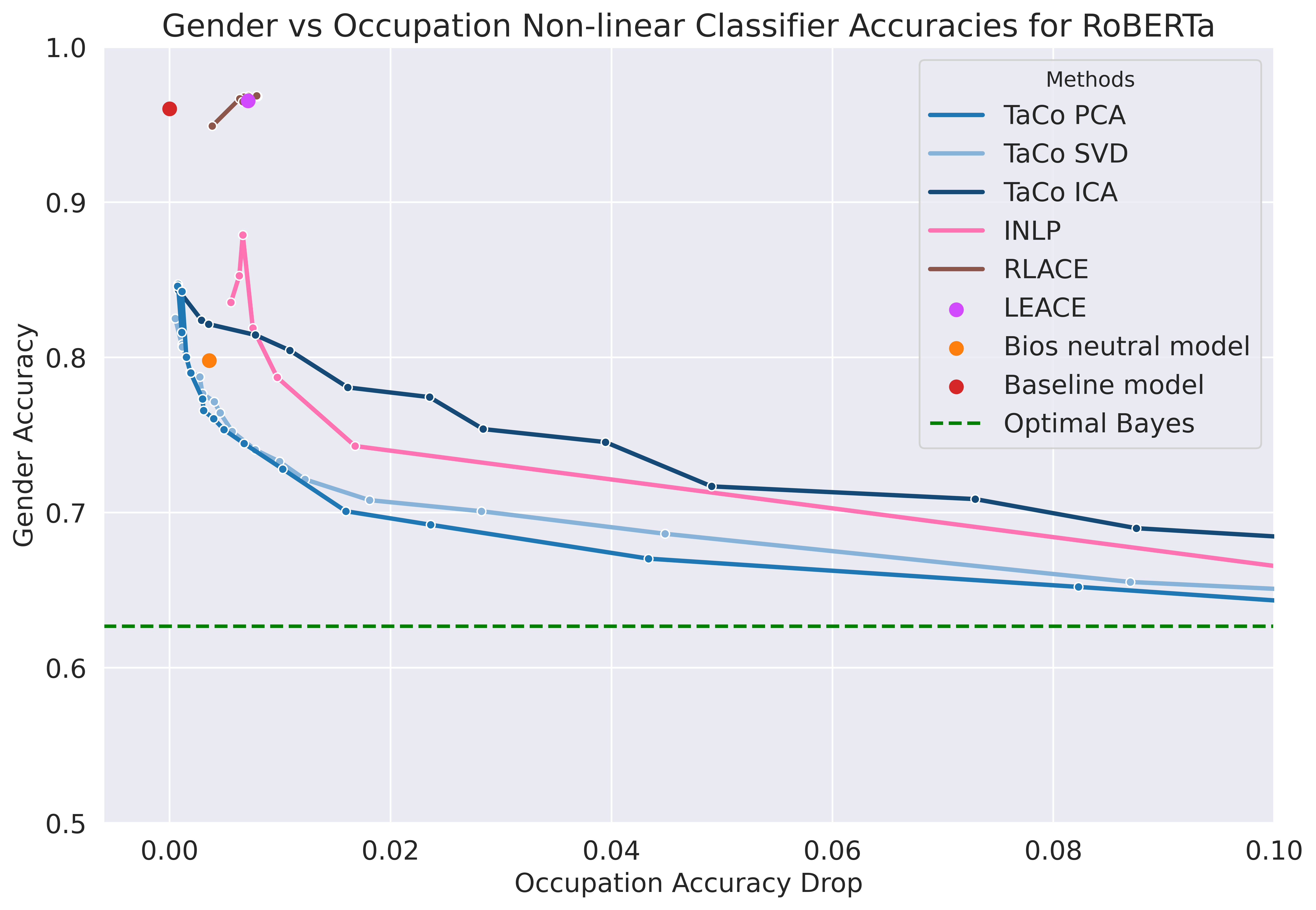}
    \end{subfigure}
    \vspace{0.5cm} 
    \begin{subfigure}{.5\textwidth}
        \centering
        \includegraphics[width=1\linewidth]{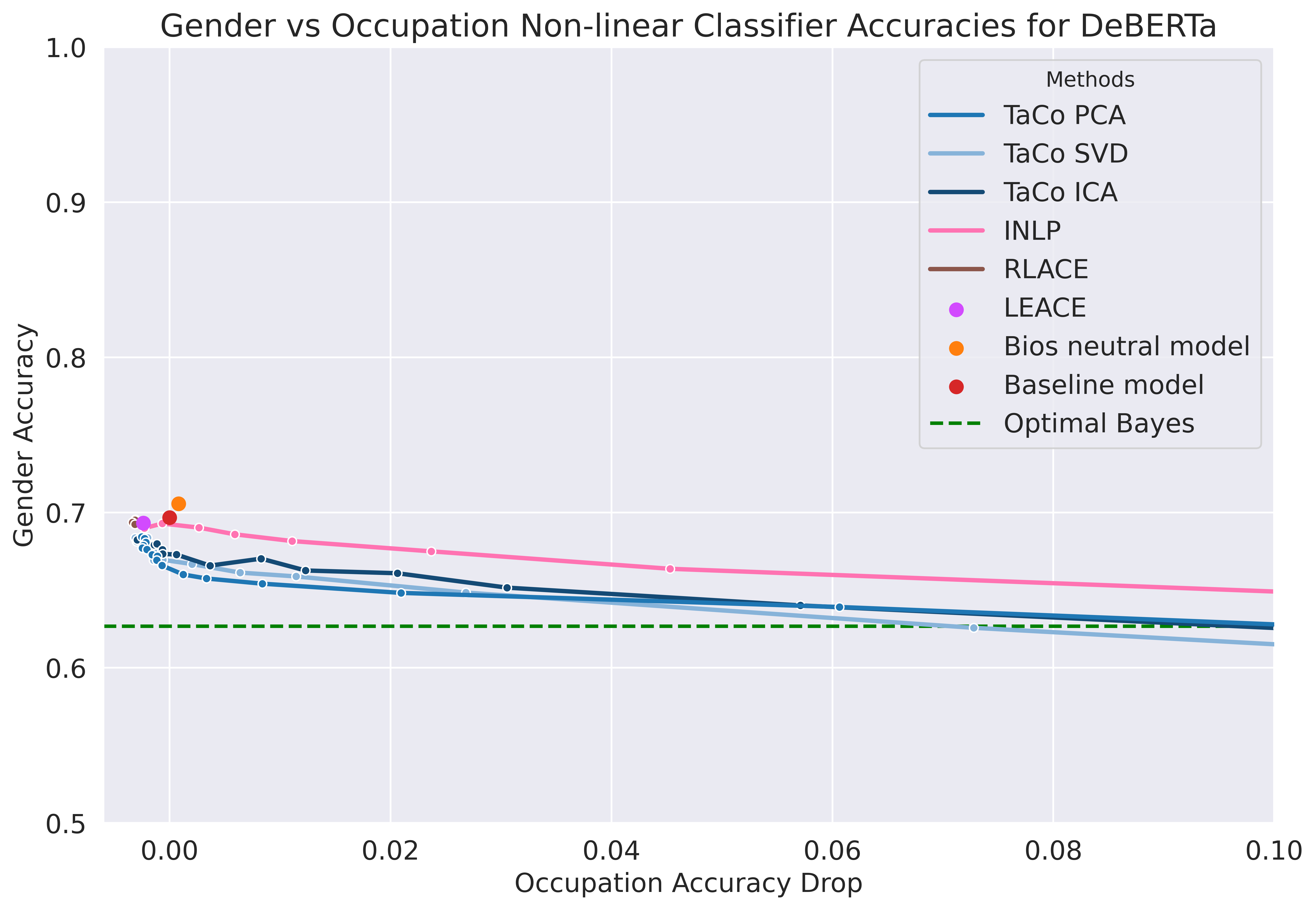}
    \end{subfigure}%
    \hspace*{\fill}
    \begin{subfigure}{.5\textwidth}
        \centering
        \includegraphics[width=1\linewidth]{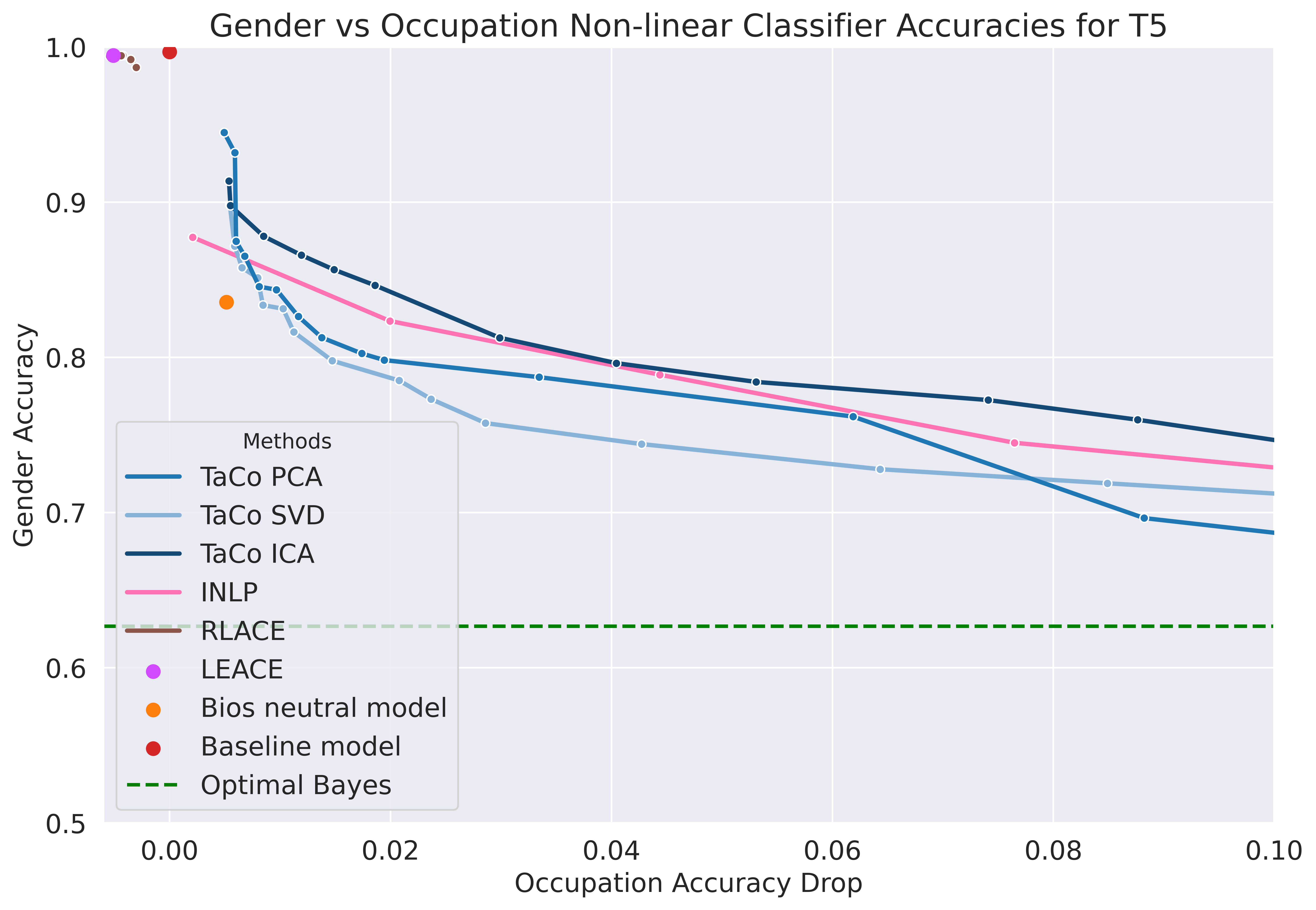}
    \end{subfigure}

    \caption{\textbf{Gender prediction accuracy versus the occupation accuracy drop -- using a two-layer MLP -- after concept erasure methods on (top left) DistilBERT, (top right) RoBERTa, (bottom left) DeBERTa, and (bottom right) T5 representations.}
    The occupation accuracy drop is reported relatively to the \texttt{Baseline model}, when the value is negative for a method, this means that occupation accuracy is better with the method than it was initially.
    For INLP and RLACE, points represent different numbers of masked dimensions.
    For TaCo methods, points represent different numbers of removed concepts. LEACE and the \textit{Bios-neutral} model are shown as single points, as they do not require parameter adjustments. The horizontal dashed line represents the accuracy of the Optimal Bayes classifier, indicating the theoretical lower bound for gender prediction accuracy.
    \textit{Figures for linear classifier on Figure \ref{apx:fig:linear_results}.}}
    \label{fig:nonlinear_results}
\end{figure*}

\section{Results}

\subsection{Dataset and models}

To evaluate the efficacy of our method, we employ the \textit{Bios} dataset \citep{de2019bias}, which comprises approximately 440,000 biographies annotated with the gender (male or female) of the individuals and their occupations across 28 classes. From this dataset, we derive a second dataset, \textit{Bios-neutral}, by removing all explicit gender indicators. This \textit{Bios-neutral} dataset serves as a baseline for comparison. The specifics of the dataset preprocessing are detailed in Appendix~\ref{apx:datasetprepro}.

For both datasets, we train three encoder transformer models: RoBERTa \citep{liu2019roberta}, DistilBERT \citep{sanh2019distilbert}, and DeBERTa \citep{he2020deberta}, and an encoder-decoder transformer model: T5 \citep{raffel2020T5}, to perform the occupation prediction task. The training procedures and hyperparameter settings for these models are provided in Appendix~\ref{apx:modelsdetailTransformers}. 

\subsection{Quantitative Evaluation of Concept Erasure}\label{sec:results-analysis-quantitative}

We compare our method against several state-of-the-art concept erasure techniques: \\
$\bullet$ Iterative Nullspace Projection (INLP) \citep{INLP} \\
$\bullet$ Reduced Linear Adversarial Concept Erasure (RLACE) \citep{RLACE} \\
$\bullet$ Linear Adversarial Concept Erasure (LEACE) \citep{LEACE}.

Additionally, we include: \\ 
$\bullet$ Baseline Model: The original model without any modifications. \\
$\bullet$ Bios-neutral Model: A model similar to the baseline but trained on the \textit{Bios-neutral} dataset. \\
$\bullet$ Optimal Bayes: corresponding to the optimal Bayes classifier $\mathbb{P}(G|Y)$ predicting gender from occupation. \\
$\bullet$ TaCo Variants: Our proposed method using different decompositions—Singular Value Decomposition (SVD), Principal Component Analysis (PCA), and Independent Component Analysis (ICA).

Figures \ref{fig:nonlinear_results} and \ref{apx:fig:linear_results} provide a comparative analysis of the debiasing methods, with the x-axis showing the drop in occupation prediction accuracy relative to the baseline model, and the y-axis representing the gender prediction accuracy. In these figures, the x-axis is truncated at a 0.1 accuracy drop, as losing more than 10\% in task accuracy is not particularly informative. The full versions of these plots, without the x-axis truncation, are provided in Figures \ref{apx:fig:nonlinear_results_all} and \ref{apx:fig:linear_results_all}, on appendix.

The varying points for each method correspond to different parameter settings: \\
$\bullet$ INLP and RLACE: We manipulate the number of dimensions masked from the final latent representation (up to 768 dimensions for BERT-type representations and 512 dimensions for T5 representation). Specifically, for INLP, we masked [200, 300, 400, 500, 550, 600, 650, 700, 750] dimensions for BERT-type representations and [100, 200, 300, 350, 400, 450, 500] dimensions for T5 representation, and for RLACE, we masked [1, 5, 10, 100, 200, 500] dimensions for BERT-type representations and [1, 5, 10, 100, 200, 300] dimensions for T5 representation. RLACE typically requires fewer dimensions to be hidden to achieve effectiveness \citep{RLACE}. \\
$\bullet$ TaCo Variants: For each decomposition (SVD, PCA, ICA), we vary the number of concepts (i.e., dimensions in the decomposition's basis) removed, ranging from 1 to $r-1$, where $r=20$ is the number of components used. This results in 19 data points for each TaCo variant. \\
$\bullet$ LEACE: Since LEACE does not require parameter tuning, it is represented by a single point. \\
$\bullet$ Baseline and Bios-neutral Models: Each is depicted by a single point corresponding to their respective accuracies. \\
$\bullet$ Optimal Bayes: Represented by a horizontal dashed line on the graph, as it has no associated occupation accuracy value. This line indicates the theoretical limit for gender prediction accuracy, below which further reduction is not meaningful.

All results are averaged over five independent runs to ensure robustness.

We assess the $\nu$-information in both linear and non-linear settings: linear $\nu$-information -- measured by the accuracy of a logistic regression classifier predicting the sensitive attribute and the label --, and non-linear $\nu$-information -- measured using a two-layer Multilayer Perceptron (MLP) classifier. Detailed results of these classifiers are provided in Appendix~\ref{apx:modelsdetailsclassif}.

As illustrated in Figure~\ref{apx:fig:linear_results}  (in Appendix), LEACE and RLACE excel at obscuring sensitive information from linear classifiers, aligning with their design objectives. Notably, TaCo with PCA closely matches INLP's performance in the linear context, despite INLP's optimization for linear scenarios.

In contrast, Figure~\ref{fig:nonlinear_results} highlights TaCo's superiority in non-linear settings. While LEACE and RLACE effectively conceal gender information from linear models, they fail to prevent non-linear models like MLPs from retrieving this information. TaCo, particularly with PCA and SVD decomposition, significantly reduces the gender information accessible to non-linear classifiers while maintaining high occupation prediction accuracy. This demonstrates a more favorable trade-off compared to simply removing explicit gender indicators, as done in the Bios-neutral model.

Further analysis reveals that TaCo with PCA slightly outperforms TaCo with SVD in non-linear contexts, especially in scenarios where task accuracy remains largely unaffected. This suggests that PCA is the optimal decomposition for TaCo when aiming to balance task performance with sensitive information removal. Conversely, TaCo with ICA performs less effectively across both linear and non-linear settings.

In summary, our proposed TaCo method, especially when combined with PCA decomposition, effectively mitigates the leakage of sensitive information in both linear and non-linear models. It achieves this with minimal impact on task performance, outperforming existing state-of-the-art techniques and offering a robust solution for bias mitigation in language models.

\subsection{Interpretability of Concepts Through Explainability Technique}

Beyond effectively erasing sensitive information from embeddings against non-linear adversaries, our method offers an additional advantage in terms of interpretability. Due to its structure, which parallels practices in explainable AI (XAI), TaCo facilitates the straightforward application of interpretability techniques to the identified concepts. Specifically, we applied the COCKATIEL method \citep{jourdan2023cockatiel} to interpret the concepts derived from our matrix decompositions before their removal. This analysis, detailed in Appendix~\ref{apx:explainability}, demonstrates that concepts obtained via SVD are semantically meaningful and provide insights into how the model processes information related to both the sensitive attribute and the primary task.  However, the interpretability of these concepts varies according to the decomposition method used, so this must be taken into account when selecting the decomposition. 

This interpretability aspect underscores the transparency of our method and highlights its potential for enhancing model accountability, an advantage that may not be as readily achievable with other concept erasure techniques.

\section{Conclusion}

In this work, we introduced TaCo, a novel concept erasure method that removes sensitive information from latent space representations, ensuring fairness against both linear and non-linear adversaries. Our method involves concept discovery, ranking based on importance to both the sensitive attribute and target label, and removing concepts that affect the sensitive variable while preserving task performance.

Experiments on the Bios dataset showed that TaCo outperforms state-of-the-art methods, effectively reducing non-linear classifiers' ability to predict sensitive attributes without compromising model accuracy.

Additionally, TaCo enhances interpretability by enabling insights into the factors influencing sensitive attribute predictions. By applying explainability techniques like COCKATIEL, we contribute to both model transparency and accountability.

Our findings highlight the importance of addressing non-linear dependencies in mitigating biases, positioning TaCo as a significant step toward fairer and more transparent AI systems.

\section*{Limitations}

While our proposed method demonstrates effectiveness in mitigating the influence of sensitive attributes, it does not guarantee a clear separation between concepts containing information about the output variable and those related to the sensitive attribute. In some cases, the concepts identified through decomposition may be inherently entangled, simultaneously encoding information pertinent to both the task and the sensitive variable. This entanglement poses a challenge in precisely isolating and removing only the unwanted information without adversely affecting the model's performance on the primary task.

Moreover, our approach to interpreting concepts post-removal has room for improvement. Interpreting concepts in natural language processing remains a complex endeavor due to the intricate and high-dimensional nature of textual data. Developing more sophisticated methods for concept interpretation is essential and warrants dedicated future research.

Another limitation pertains to the scope of our application. While we focused on mitigating gender bias using the \textit{Bios} dataset, which considers gender as a binary attribute, we acknowledge that gender is inherently non-binary. Consequently, models trained on such datasets may not fully capture the spectrum of gender identities present in real-world scenarios. Although our method is designed to operate effectively with variables encompassing multiple classes, including non-binary gender categorizations, the lack of datasets reflecting this diversity limits the evaluation of our approach in more inclusive settings. This underscores the need for the NLP community to develop and adopt datasets that incorporate non-binary and diverse representations of sensitive attributes, thereby promoting fairness and inclusivity in AI applications.


\section*{Ethics Statement}
With the advent of LLMs and their expanding applications, it is crucial to ensure that biases are adequately addressed before deploying these models in high-stakes applications. In this paper, we introduce a method that not only mitigates biases but does so in a manner that promotes human understanding of the specific biases removed from the model. Transparency in identifying bias sources and understanding the trade-off involved are vital aspects of our approach.

An important consideration is the computational efficiency of our method, which significantly reduces the resource burden. Our approach only requires training a classification head while keeping the backbone of the model unchanged. This computational affordability makes our method particularly valuable for companies with limited budgets, enabling them to effectively address bias concerns without incurring excessive costs.

We believe that it is our ethical responsibility to prioritize fairness and accountability in the development and deployment of AI systems. By providing a debiasing technique that offers interpretability, computational efficiency, cost-effectiveness and ease of use, we aim to support the responsible use of LLMs, ensuring they are better aligned with ethical standards and advancing the progress of unbiased AI technologies.

\section*{Acknowledgements}
We thank the ANR-3IA Artificial and Natural Intelligence Toulouse Institute (ANITI) funded by the ANR-19-PI3A-0004 grant for research support.
This work was conducted as part of the DEEL\footnote{\url{https://www.deel.ai}} project.


\bibliography{custom,more-refs-from-nicholas}

\begin{thebibliography}{46}
\expandafter\ifx\csname natexlab\endcsname\relax\def\natexlab#1{#1}\fi

\bibitem[{Asher et~al.(2022)Asher, De~Lara, Paul, and Russell}]{asher:etal:2022}
Nicholas Asher, Lucas De~Lara, Soumya Paul, and Chris Russell. 2022.
\newblock Counterfactual models for fair and adequate explanations.
\newblock \emph{Machine Learning and Knowledge Extraction}, 4(2):316--349.

\bibitem[{Belrose et~al.(2023)Belrose, Schneider-Joseph, Ravfogel, Cotterell, Raff, and Biderman}]{LEACE}
Nora Belrose, David Schneider-Joseph, Shauli Ravfogel, Ryan Cotterell, Edward Raff, and Stella Biderman. 2023.
\newblock Leace: Perfect linear concept erasure in closed form.
\newblock \emph{arXiv preprint arXiv:2306.03819}.

\bibitem[{Bolukbasi et~al.(2016)Bolukbasi, Chang, Zou, Saligrama, and Kalai}]{bolukbasi2016man}
Tolga Bolukbasi, Kai-Wei Chang, James~Y Zou, Venkatesh Saligrama, and Adam~T Kalai. 2016.
\newblock Man is to computer programmer as woman is to homemaker? debiasing word embeddings.
\newblock \emph{Advances in neural information processing systems}, 29.

\bibitem[{Caliskan et~al.(2017)Caliskan, Bryson, and Narayanan}]{caliskan2017semantics}
Aylin Caliskan, Joanna~J Bryson, and Arvind Narayanan. 2017.
\newblock Semantics derived automatically from language corpora contain human-like biases.
\newblock \emph{Science}, 356(6334):183--186.

\bibitem[{Comon(1994)}]{comon1994independent}
Pierre Comon. 1994.
\newblock Independent component analysis, a new concept?
\newblock \emph{Signal processing}, 36(3):287--314.

\bibitem[{De-Arteaga et~al.(2019)De-Arteaga, Romanov, Wallach, Chayes, Borgs, Chouldechova, Geyik, Kenthapadi, and Kalai}]{de2019bias}
Maria De-Arteaga, Alexey Romanov, Hanna Wallach, Jennifer Chayes, Christian Borgs, Alexandra Chouldechova, Sahin Geyik, Krishnaram Kenthapadi, and Adam~Tauman Kalai. 2019.
\newblock Bias in bios: A case study of semantic representation bias in a high-stakes setting.
\newblock In \emph{proceedings of the Conference on Fairness, Accountability, and Transparency}, pages 120--128.

\bibitem[{Devlin et~al.(2018)Devlin, Chang, Lee, and Toutanova}]{devlin2018bert}
Jacob Devlin, Ming-Wei Chang, Kenton Lee, and Kristina Toutanova. 2018.
\newblock Bert: Pre-training of deep bidirectional transformers for language understanding.
\newblock \emph{arXiv preprint arXiv:1810.04805}.

\bibitem[{Eckart and Young(1936)}]{EckartYoung1936}
C.~Eckart and G.~Young. 1936.
\newblock The approximation of one matrix by another of lower rank.
\newblock \emph{Psychometrika}, 1:211--218.

\bibitem[{Fel et~al.(2023)Fel, Picard, Bethune, Boissin, Vigouroux, Colin, Cad{\`e}ne, and Serre}]{fel:etal:2023}
Thomas Fel, Agustin Picard, Louis Bethune, Thibaut Boissin, David Vigouroux, Julien Colin, R{\'e}mi Cad{\`e}ne, and Thomas Serre. 2023.
\newblock Craft: Concept recursive activation factorization for explainability.
\newblock In \emph{Proceedings of the IEEE/CVF Conference on Computer Vision and Pattern Recognition}, pages 2711--2721.

\bibitem[{Field and Tsvetkov(2020)}]{field2020unsupervised}
Anjalie Field and Yulia Tsvetkov. 2020.
\newblock Unsupervised discovery of implicit gender bias.
\newblock \emph{arXiv preprint arXiv:2004.08361}.

\bibitem[{Garg et~al.(2018)Garg, Schiebinger, Jurafsky, and Zou}]{garg2018word}
Nikhil Garg, Londa Schiebinger, Dan Jurafsky, and James Zou. 2018.
\newblock Word embeddings quantify 100 years of gender and ethnic stereotypes.
\newblock \emph{Proceedings of the National Academy of Sciences}, 115(16):E3635--E3644.

\bibitem[{Gonen and Goldberg(2019)}]{gonen2019lipstick}
Hila Gonen and Yoav Goldberg. 2019.
\newblock Lipstick on a pig: Debiasing methods cover up systematic gender biases in word embeddings but do not remove them.
\newblock \emph{arXiv preprint arXiv:1903.03862}.

\bibitem[{He et~al.(2020)He, Liu, Gao, and Chen}]{he2020deberta}
Pengcheng He, Xiaodong Liu, Jianfeng Gao, and Weizhu Chen. 2020.
\newblock Deberta: Decoding-enhanced bert with disentangled attention.
\newblock \emph{arXiv preprint arXiv:2006.03654}.

\bibitem[{Jacovi and Goldberg(2020)}]{jacovi:goldberg:2020}
Alon Jacovi and Yoav Goldberg. 2020.
\newblock Towards faithfully interpretable nlp systems: How should we define and evaluate faithfulness?
\newblock In \emph{Proceedings of the 58th Annual Meeting of the Association for Computational Linguistics}, pages 4198--4205.

\bibitem[{Jacovi et~al.(2021)Jacovi, Swayamdipta, Ravfogel, Elazar, Choi, and Goldberg}]{jacovi:etal:2021}
Alon Jacovi, Swabha Swayamdipta, Shauli Ravfogel, Yanai Elazar, Yejin Choi, and Yoav Goldberg. 2021.
\newblock Contrastive explanations for model interpretability.
\newblock In \emph{Proceedings of the 2021 Conference on Empirical Methods in Natural Language Processing}, pages 1597--1611.

\bibitem[{Janon et~al.(2014)Janon, Klein, Lagnoux, Nodet, and Prieur}]{janon2014asymptotic}
Alexandre Janon, Thierry Klein, Agnes Lagnoux, Ma{\"e}lle Nodet, and Cl{\'e}mentine Prieur. 2014.
\newblock Asymptotic normality and efficiency of two sobol index estimators.
\newblock \emph{ESAIM: Probability and Statistics}, 18:342--364.

\bibitem[{Jourdan et~al.(2023)Jourdan, Picard, Fel, Risser, Loubes, and Asher}]{jourdan2023cockatiel}
Fanny Jourdan, Agustin Picard, Thomas Fel, Laurent Risser, Jean~Michel Loubes, and Nicholas Asher. 2023.
\newblock Cockatiel: Continuous concept ranked attribution with interpretable elements for explaining neural net classifiers on nlp tasks.
\newblock \emph{in Proceedings of the Findings of the Association for Computational Linguistics (ACL 2023).}

\bibitem[{Karita et~al.(2019)Karita, Chen, Hayashi, Hori, Inaguma, Jiang, Someki, Soplin, Yamamoto, Wang et~al.}]{karita2019comparative}
Shigeki Karita, Nanxin Chen, Tomoki Hayashi, Takaaki Hori, Hirofumi Inaguma, Ziyan Jiang, Masao Someki, Nelson Enrique~Yalta Soplin, Ryuichi Yamamoto, Xiaofei Wang, et~al. 2019.
\newblock A comparative study on transformer vs rnn in speech applications.
\newblock In \emph{2019 IEEE Automatic Speech Recognition and Understanding Workshop (ASRU)}, pages 449--456. IEEE.

\bibitem[{Kilbertus et~al.(2017)Kilbertus, Rojas~Carulla, Parascandolo, Hardt, Janzing, and Sch{\"o}lkopf}]{kilbertus2017avoiding}
Niki Kilbertus, Mateo Rojas~Carulla, Giambattista Parascandolo, Moritz Hardt, Dominik Janzing, and Bernhard Sch{\"o}lkopf. 2017.
\newblock Avoiding discrimination through causal reasoning.
\newblock \emph{Advances in neural information processing systems}, 30.

\bibitem[{Kim et~al.(2018)Kim, Wattenberg, Gilmer, Cai, Wexler, Viegas et~al.}]{kim2018interpretability}
Been Kim, Martin Wattenberg, Justin Gilmer, Carrie Cai, James Wexler, Fernanda Viegas, et~al. 2018.
\newblock Interpretability beyond feature attribution: Quantitative testing with concept activation vectors (tcav).
\newblock In \emph{International conference on machine learning}, pages 2668--2677. PMLR.

\bibitem[{Lewis(1973)}]{lewis:1973}
David Lewis. 1973.
\newblock \emph{Counterfactuals}.
\newblock Basil Blackwell, Oxford.

\bibitem[{Liang et~al.(2020)Liang, Li, Zheng, Lim, Salakhutdinov, and Morency}]{liang2020towards}
Paul~Pu Liang, Irene~Mengze Li, Emily Zheng, Yao~Chong Lim, Ruslan Salakhutdinov, and Louis-Philippe Morency. 2020.
\newblock Towards debiasing sentence representations.
\newblock \emph{arXiv preprint arXiv:2007.08100}.

\bibitem[{Liu et~al.(2019{\natexlab{a}})Liu, Ott, Goyal, Du, Joshi, Chen, Levy, Lewis, Zettlemoyer, and Stoyanov}]{liu2019roberta}
Yinhan Liu, Myle Ott, Naman Goyal, Jingfei Du, Mandar Joshi, Danqi Chen, Omer Levy, Mike Lewis, Luke Zettlemoyer, and Veselin Stoyanov. 2019{\natexlab{a}}.
\newblock Roberta: A robustly optimized bert pretraining approach.
\newblock \emph{arXiv preprint arXiv:1907.11692}.

\bibitem[{Liu et~al.(2019{\natexlab{b}})Liu, Ott, Goyal, Du, Joshi, Chen, Levy, Lewis, Zettlemoyer, and Stoyanov}]{DBLP:journals/corr/abs-1907-11692}
Yinhan Liu, Myle Ott, Naman Goyal, Jingfei Du, Mandar Joshi, Danqi Chen, Omer Levy, Mike Lewis, Luke Zettlemoyer, and Veselin Stoyanov. 2019{\natexlab{b}}.
\newblock \href {http://arxiv.org/abs/1907.11692} {Roberta: {A} robustly optimized {BERT} pretraining approach}.
\newblock \emph{CoRR}, abs/1907.11692.

\bibitem[{Mackenzie et~al.(2020)Mackenzie, Benham, Petri, Trippas, Culpepper, and Moffat}]{10.1145/3340531.3412762}
Joel Mackenzie, Rodger Benham, Matthias Petri, Johanne~R. Trippas, J.~Shane Culpepper, and Alistair Moffat. 2020.
\newblock \href {https://doi.org/10.1145/3340531.3412762} {Cc-news-en: A large english news corpus}.
\newblock In \emph{Proceedings of the 29th ACM International Conference on Information amp; Knowledge Management}, CIKM '20, page 3077–3084, New York, NY, USA. Association for Computing Machinery.

\bibitem[{Marrel et~al.(2009)Marrel, Iooss, Laurent, and Roustant}]{marrel2009calculations}
Amandine Marrel, Bertrand Iooss, Beatrice Laurent, and Olivier Roustant. 2009.
\newblock Calculations of sobol indices for the gaussian process metamodel.
\newblock \emph{Reliability Engineering \& System Safety}, 94(3):742--751.

\bibitem[{Pearl(2009)}]{pearl2009causality}
Judea Pearl. 2009.
\newblock \emph{Causality}.
\newblock Cambridge university press.

\bibitem[{Radford et~al.(2019)Radford, Wu, Child, Luan, Amodei, Sutskever et~al.}]{radford2019language}
Alec Radford, Jeffrey Wu, Rewon Child, David Luan, Dario Amodei, Ilya Sutskever, et~al. 2019.
\newblock Language models are unsupervised multitask learners.
\newblock \emph{OpenAI blog}, 1(8):9.

\bibitem[{Raffel et~al.(2020)Raffel, Shazeer, Roberts, Lee, Narang, Matena, Zhou, Li, and Liu}]{raffel2020T5}
Colin Raffel, Noam Shazeer, Adam Roberts, Katherine Lee, Sharan Narang, Michael Matena, Yanqi Zhou, Wei Li, and Peter~J Liu. 2020.
\newblock Exploring the limits of transfer learning with a unified text-to-text transformer.
\newblock \emph{Journal of machine learning research}, 21(140):1--67.

\bibitem[{Ravfogel et~al.(2020)Ravfogel, Elazar, Gonen, Twiton, and Goldberg}]{INLP}
Shauli Ravfogel, Yanai Elazar, Hila Gonen, Michael Twiton, and Yoav Goldberg. 2020.
\newblock Null it out: Guarding protected attributes by iterative nullspace projection.
\newblock \emph{arXiv preprint arXiv:2004.07667}.

\bibitem[{Ravfogel et~al.(2023)Ravfogel, Goldberg, and Cotterell}]{ravfogel2023loglinear}
Shauli Ravfogel, Yoav Goldberg, and Ryan Cotterell. 2023.
\newblock Log-linear guardedness and its implications.
\newblock In \emph{Proceedings of the 61st Annual Meeting of the Association for Computational Linguistics (Volume 1: Long Papers)}, pages 9413--9431.

\bibitem[{Ravfogel et~al.(2022)Ravfogel, Twiton, Goldberg, and Cotterell}]{RLACE}
Shauli Ravfogel, Michael Twiton, Yoav Goldberg, and Ryan~D Cotterell. 2022.
\newblock Linear adversarial concept erasure.
\newblock In \emph{International Conference on Machine Learning}, pages 18400--18421. PMLR.

\bibitem[{Saltelli et~al.(2010)Saltelli, Annoni, Azzini, Campolongo, Ratto, and Tarantola}]{saltelli2010variance}
Andrea Saltelli, Paola Annoni, Ivano Azzini, Francesca Campolongo, Marco Ratto, and Stefano Tarantola. 2010.
\newblock Variance based sensitivity analysis of model output. design and estimator for the total sensitivity index.
\newblock \emph{Computer physics communications}, 181(2):259--270.

\bibitem[{Sanh et~al.(2019)Sanh, Debut, Chaumond, and Wolf}]{sanh2019distilbert}
Victor Sanh, Lysandre Debut, Julien Chaumond, and Thomas Wolf. 2019.
\newblock Distilbert, a distilled version of bert: smaller, faster, cheaper and lighter.
\newblock \emph{arXiv preprint arXiv:1910.01108}.

\bibitem[{Shao et~al.(2023)Shao, Ziser, and Cohen}]{shao2023erasure}
Shun Shao, Yftah Ziser, and Shay~B Cohen. 2023.
\newblock Erasure of unaligned attributes from neural representations.
\newblock \emph{Transactions of the Association for Computational Linguistics}, 11:488--510.

\bibitem[{Sobol(1993)}]{sobol1993sensitivity}
Ilya~M Sobol. 1993.
\newblock Sensitivity analysis for non-linear mathematical models.
\newblock \emph{Mathematical modelling and computational experiment}, 1:407--414.

\bibitem[{Trinh and Le(2018)}]{trinh2018simple}
Trieu~H Trinh and Quoc~V Le. 2018.
\newblock A simple method for commonsense reasoning.
\newblock \emph{arXiv preprint arXiv:1806.02847}.

\bibitem[{Wiegreffe and Pinter(2019)}]{wiegreffe:pinter:2019}
Sarah Wiegreffe and Yuval Pinter. 2019.
\newblock Attention is not not explanation.
\newblock In \emph{Proceedings of the 2019 Conference on Empirical Methods in Natural Language Processing and the 9th International Joint Conference on Natural Language Processing (EMNLP-IJCNLP)}, pages 11--20.

\bibitem[{Wold et~al.(1987)Wold, Esbensen, and Geladi}]{wold1987principal}
Svante Wold, Kim Esbensen, and Paul Geladi. 1987.
\newblock Principal component analysis.
\newblock \emph{Chemometrics and intelligent laboratory systems}, 2(1-3):37--52.

\bibitem[{Xu et~al.(2020)Xu, Zhao, Song, Stewart, and Ermon}]{Xu2020A}
Yilun Xu, Shengjia Zhao, Jiaming Song, Russell Stewart, and Stefano Ermon. 2020.
\newblock \href {https://openreview.net/forum?id=r1eBeyHFDH} {A theory of usable information under computational constraints}.
\newblock In \emph{International Conference on Learning Representations}.

\bibitem[{Yin and Neubig(2022)}]{yin:neubig:2022}
Kayo Yin and Graham Neubig. 2022.
\newblock Interpreting language models with contrastive explanations.
\newblock In \emph{Proceedings of the 2022 Conference on Empirical Methods in Natural Language Processing}, pages 184--198.

\bibitem[{Zeiler and Fergus(2014)}]{zeiler2014visualizing}
Matthew~D Zeiler and Rob Fergus. 2014.
\newblock Visualizing and understanding convolutional networks.
\newblock In \emph{European conference on computer vision}, pages 818--833. Springer.

\bibitem[{Zhao et~al.(2019)Zhao, Wang, Yatskar, Cotterell, Ordonez, and Chang}]{zhao2019gender}
Jieyu Zhao, Tianlu Wang, Mark Yatskar, Ryan Cotterell, Vicente Ordonez, and Kai-Wei Chang. 2019.
\newblock Gender bias in contextualized word embeddings.
\newblock \emph{arXiv preprint arXiv:1904.03310}.

\bibitem[{Zhao et~al.(2018{\natexlab{a}})Zhao, Wang, Yatskar, Ordonez, and Chang}]{zhao2018gender}
Jieyu Zhao, Tianlu Wang, Mark Yatskar, Vicente Ordonez, and Kai-Wei Chang. 2018{\natexlab{a}}.
\newblock Gender bias in coreference resolution: Evaluation and debiasing methods.
\newblock \emph{arXiv preprint arXiv:1804.06876}.

\bibitem[{Zhao et~al.(2018{\natexlab{b}})Zhao, Zhou, Li, Wang, and Chang}]{zhao2018learning}
Jieyu Zhao, Yichao Zhou, Zeyu Li, Wei Wang, and Kai-Wei Chang. 2018{\natexlab{b}}.
\newblock Learning gender-neutral word embeddings.
\newblock \emph{arXiv preprint arXiv:1809.01496}.

\bibitem[{Zhu et~al.(2015)Zhu, Kiros, Zemel, Salakhutdinov, Urtasun, Torralba, and Fidler}]{moviebook}
Yukun Zhu, Ryan Kiros, Richard Zemel, Ruslan Salakhutdinov, Raquel Urtasun, Antonio Torralba, and Sanja Fidler. 2015.
\newblock Aligning books and movies: Towards story-like visual explanations by watching movies and reading books.
\newblock In \emph{arXiv preprint arXiv:1506.06724}.

\end{thebibliography}
\bibliographystyle{acl_natbib}

\appendix

\section{Gender prediction with RoBERTa}\label{apx:genderpredrob}

In this section, we study the impact on gender prediction of removing explicit gender indicators with a transformer model. To do this, we train a model on a dataset and we train a model with the same architecture on modified version of the dataset where we remove the explicit gender indicators. We train these models to predict the gender and we compare both of these model's accuracies.

In our example, we use the \textit{Bios} dataset \cite{de2019bias}, which contains about 400K biographies (textual data). For each biography, we have the gender (M or F) associated. We use each biography to predict the gender, and then we clean the dataset and we create a second dataset without the explicit gender indicators (following the protocol in Appendix \ref{apx:datasetprepro}).

We train a RoBERTa model, as explained in Appendix \ref{apx:modelsdetailTransformers}. 
Instead of training the model on the occupations, we place ourselves directly in the case where we train the RoBERTa to predict gender (with the same parameters). We train both models over 5 epochs.

\begin{figure}[ht]
    \centering
\includegraphics[width=1.0\linewidth]{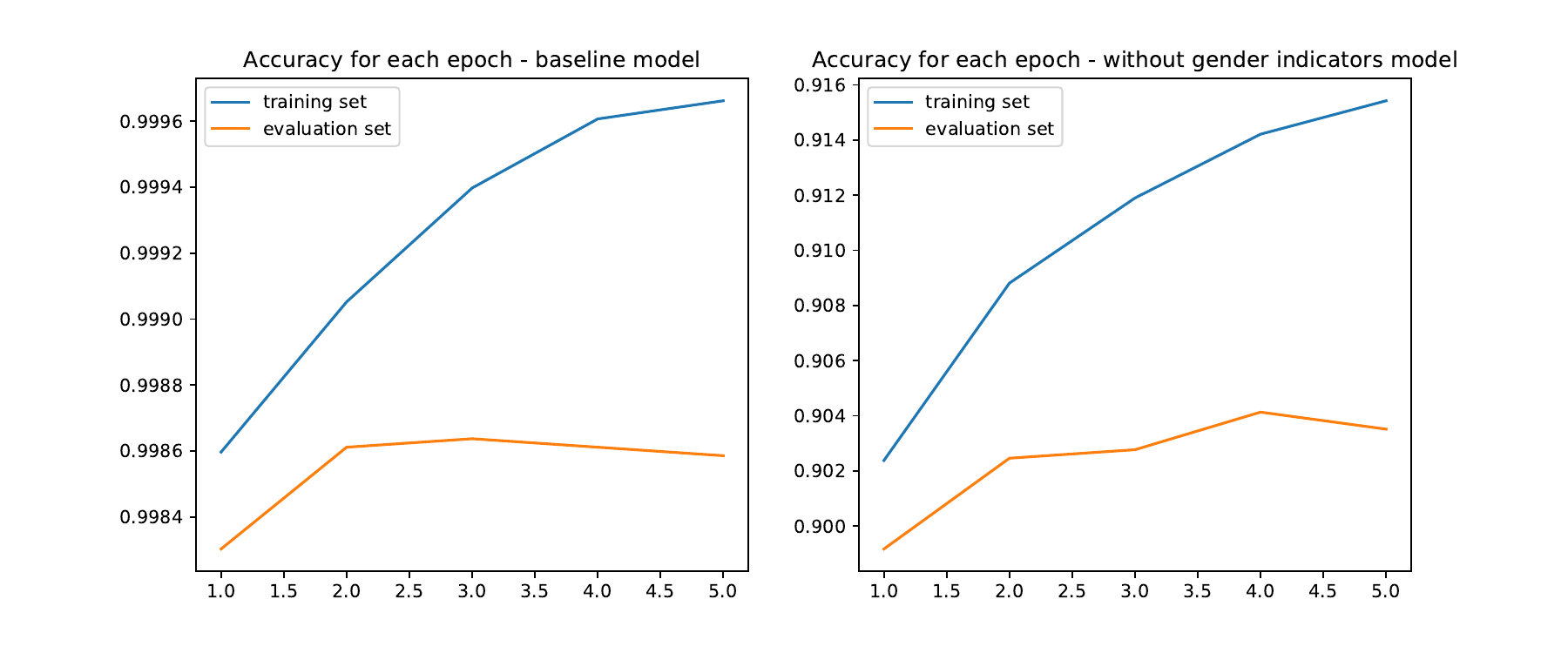}
    \caption{Convergence curves (accuracy) during training for the gender classification task, 5 epochs. (On the left) RoBERTa model trained on \textit{Bios}; (On the right): RoBERTa model trained on \textit{Bios} without explicit gender indicators.}
    \label{fig:acccurves}
\end{figure}

In figure \ref{fig:acccurves}, we go from 99\% accuracy for the baseline to 90\% accuracy for the model trained on the dataset without explicit gender indicators. Even with a dataset without explicit gender indicators, a model like RoBERTa performs well on the gender prediction task. 

\section{Dataset pre-processing protocol}\label{apx:datasetprepro}

We use \textit{Bios} dataset \cite{de2019bias}, which contains about 400K biographies (textual data). For each biography, \textit{Bios} specifies the gender (M or F) of its author as well as its occupation (among a total of 28 possible occupations, categorical data).

\begin{figure}[ht]
    \centering
\includegraphics[width=1.0\linewidth]{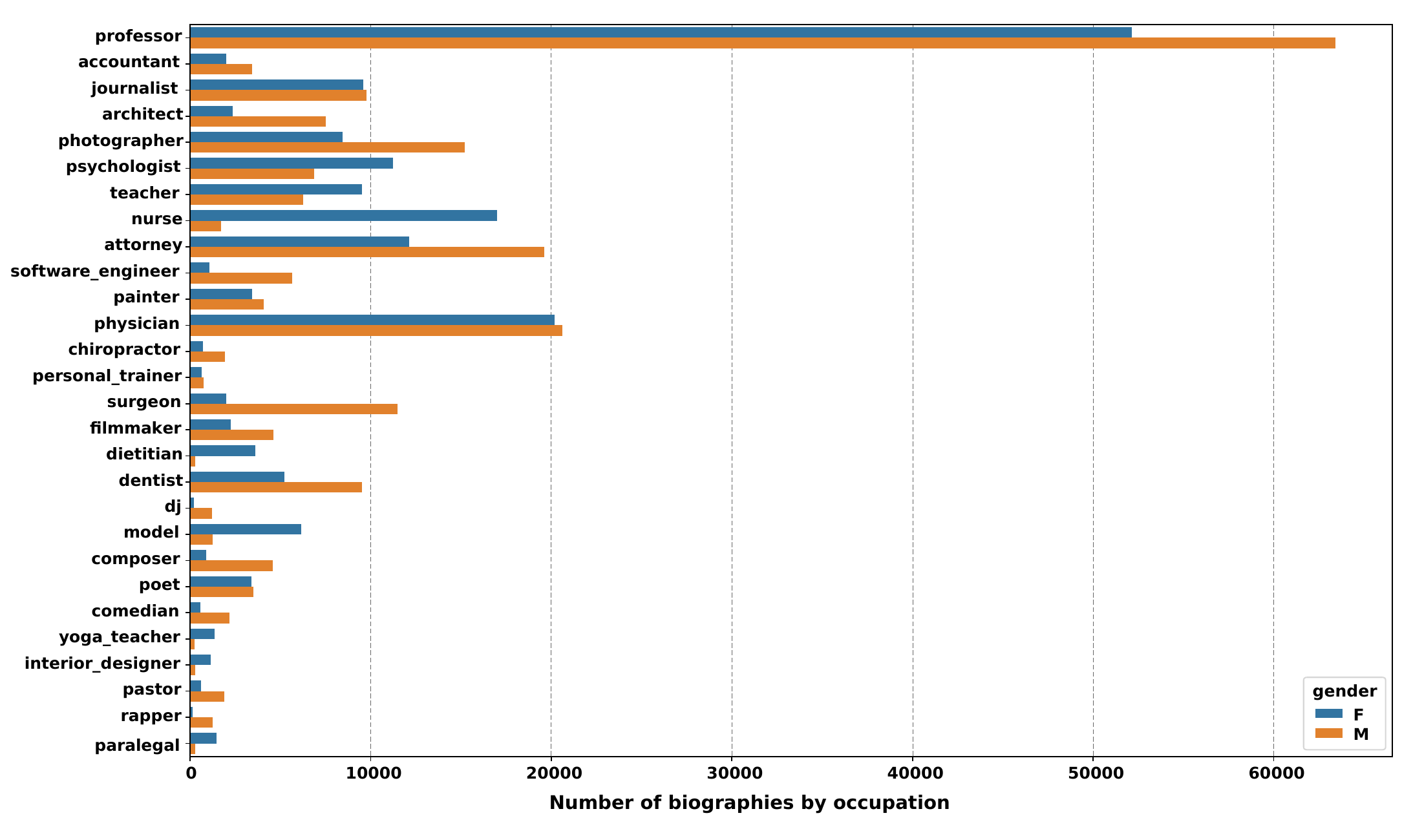}
    \caption{Number of biographies for each occupation by gender on the total \textit{Bios} dataset \cite{de2019bias}.}
    \label{fig:distribution}
\end{figure}

\begin{figure}
    \centering
\includegraphics[width=0.8\linewidth]{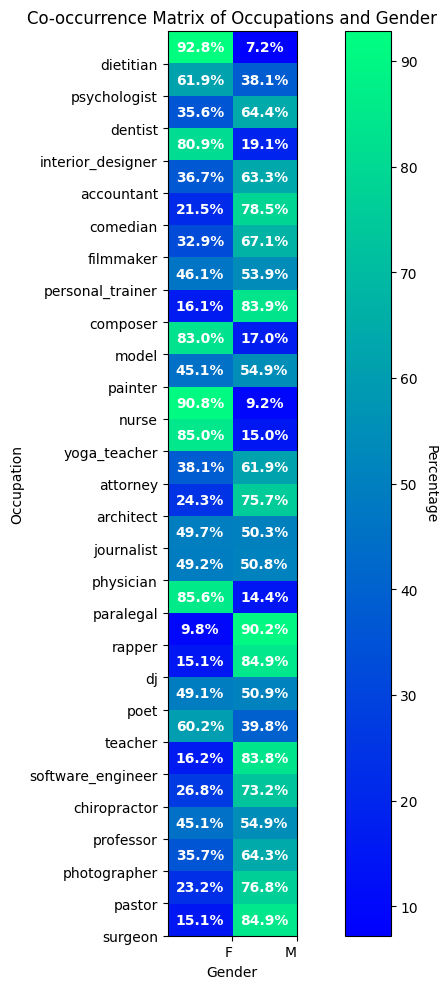}
    \caption{Co-occurrence of occupation and gender for \textit{Bios} dataset \cite{de2019bias}. Solely predicting the gender from the occupation yields a minimum of 62\% accuracy, which is the minimum baseline for an accurate but unbiased model. Any classifier with lower accuracy on gender prediction must have accuracy on occupation prediction lower than $100\%$.}
    \label{fig:cooccur}
\end{figure}

As shown in Figure~\ref{fig:distribution}, this dataset contains heterogeneously represented occupations. Although the representation of some occupations is relatively well-balanced between males and females, other occupations are particularly unbalanced (see Figure \ref{fig:cooccur}).

\paragraph{Cleaning dataset}
To clean the dataset, we take all biographies and apply these modifications: \vspace{-2mm}
\begin{itemize}
    \item we remove mail and URL, \vspace{-2mm}
    \item we remove dot after pronouns and acronyms, \vspace{-2mm}
    \item we remove double "?", "!" and ".", \vspace{-2mm}
    \item we cut all the biographies at 512 tokens by checking that they are cut at the end of a sentence and not in the middle.
\end{itemize}

\paragraph{Creating the \textit{Bios-neutral} dataset}

From \textit{Bios}, we create a new dataset without explicit gender indicators, called\textit{ Bios-neutral}. 

At first, we tokenize the dataset, then: \vspace{-2mm} \begin{itemize}
    \item If the token is a first name, we replace it with "\textit{Sam}" (a common neutral first name). \vspace{-2mm}
    \item If the token is a dictionary key (on the dictionary defined below), we replace it with its values. \vspace{-2mm}
\end{itemize}

To determine whether the token is a first name, we use the list of first names:  \href{https://www.usna.edu/Users/cs/roche/courses/s15si335/proj1/files.php\%3Ff=names.txt&downloadcode=yes}{usna.edu}

The dictionary used in the second part is based on the one created in \cite{field2020unsupervised}.

\section{Models details and convergence curves}\label{apx:modelsdetails}

\subsection{Transformer training}\label{apx:modelsdetailTransformers}
\paragraph{RoBERTa base training}
We use a RoBERTa model \citep{liu2019roberta}, which is based on the transformers architecture and is pre-trained with the Masked language modeling (MLM) objective. 
We specifically used a RoBERTa base model pre-trained by HuggingFace. All information related to how it was trained can be found in  \citet{DBLP:journals/corr/abs-1907-11692}. It can be remarked, that a very large training dataset was used to pre-train the model, as it was composed of five datasets: \emph{BookCorpus} \citep{moviebook}, a dataset containing 11,038 unpublished books; \emph{English Wikipedia} (excluding lists, tables and headers); \emph{CC-News} \citep{10.1145/3340531.3412762} which contains 63 millions English news articles crawled between September 2016 and February 2019; \emph{OpenWebText} \citep{radford2019language}  an open-source recreation of the WebText dataset used to train GPT-2; \emph{Stories} \citep{trinh2018simple} a dataset containing a subset of CommonCrawl data filtered to match the story-like style of Winograd schemas. Pre-training was performed on these data by randomly masking 15\% of the words in each of the input sentences and then trying to predict the masked words.

\paragraph{DistilBERT base training}
DistilBERT \citep{sanh2019distilbert} is a transformer architecture derivative from but smaller and faster than the original BERT \citep{devlin2018bert}. 
This model is commonly used to do text classification. DistilBERT is trained on BookCorpus \cite{moviebook} (like BERT), a dataset consisting of 11,038 unpublished books and English Wikipedia (excluding lists, tables and headers), using the BERT base model as a teacher.

\paragraph{DeBERTa base training}
DeBERTa \citep{he2020deberta} enhances the BERT and RoBERTa models by incorporating disentangled attention and an improved mask decoder. These enhancements enable DeBERTa to surpass RoBERTa in most Natural Language Understanding (NLU) tasks, utilizing 80GB of training data. In the DeBERTa V3 iteration (used here),  DeBERTa is optimized efficiently through ELECTRA-Style pre-training and Gradient Disentangled Embedding Sharing. This version, compared to its predecessor, shows significant performance gains in downstream tasks. 

The base model of DeBERTa V3 features 12 layers with a hidden size of 768. It possesses 86M core parameters, and its vocabulary includes 128K tokens, adding 98M parameters to the Embedding layer. This model was trained on 160GB of data, similar to DeBERTa V2.

\paragraph{T5-small base training}
The T5 model \citep{raffel2020T5}, short for "Text-To-Text Transfer Transformer," follows a unified framework where all NLP tasks are cast as a text-to-text format, with both input and output represented as strings. This model, pre-trained by Hugging Face, is based on the transformer architecture and uses a sequence-to-sequence structure. T5-small, a more lightweight version of the full model, contains 60 million parameters and is pre-trained on the C4 dataset, a massive corpus extracted from publicly available web pages. The training objective involved a masked span prediction task, where consecutive tokens in the input are masked, and the model is trained to reconstruct the missing spans. This architecture and training objective allow T5 to perform a wide range of downstream tasks, including text classification, summarization, and translation, by leveraging its ability to convert any task into a text generation problem.

\paragraph{Occupation prediction task}
After pre-training RoBERTa/DistilBERT/DeBERTa/T5 parameters on this huge dataset, we then trained it on the 400.000 biographies of the  \emph{Bios} dataset. 
The training was performed with PyTorch on 2 GPUs (Nvidia Quadro RTX6000 24GB RAM) for 10/3/3/2 epochs (we train 15 epochs, but we perform early stopping, giving the best results at 10/3/3/2 epochs) with a batch size of 8 observations and a sequence length of 512 words. The optimizer was Adam with a learning rate of 1e-6, $\beta_1 = 0.9$, $\beta_2 = 0.98$, and $\epsilon =1e6$. 

We split the dataset into 70\% for training, 10\% for validation, and 20\% for testing. 

\subsection{Classifier heads training}\label{apx:modelsdetailsclassif}

\subsubsection{Non-linear classifiers}
The importance estimation in part 2 of our method requires a classifier to be trained on top of the features. An ideal classifier will leverage the maximum amount possible of information about gender or occupation that can be realistically extracted from latent features, in the spirit of the maximum information that can be extracted under computational constraints~\citep{Xu2020A}. Hence, it is crucial to optimize an expressive model to the highest possible performance.  

In the latent space of the feature extractor, the decisions are usually computed with a linear classifier. However, when operating on latent space with removed concepts or when working with a task on which the transformer head has not been fine-tuned, nothing guarantees that the task can be solved with linear probes. Hence, we chose a more expressive model, namely a two-layer perceptron with ReLU non-linearities, and architecture $768\rightarrow 128\rightarrow C$ with $C$ the number of classes. The network is trained by minimizing the categorical cross-entropy on the \textit{train set}, with \textit{Softmax} activation to convert the logits into probabilities. We use Adam with a learning rate chosen between $10^{-5}$ and $10^{-1}$ with cross-validation on the \textit{validation set}. Note that the paper reports the accuracy on the \textit{test set}.  

The gender classifier $c_{g}$ and the occupation classifier $\tilde{c}$ are both trained with this same protocol.

\subsubsection{Linear classifiers}
To compare with state-of-the-art models, we also train simple logistic regressions for the gender classifier $c_g$ and the occupation classifier $\tilde{c}$. We use \texttt{linear\_model.LogisticRegression} from the \texttt{scikit-learn} package. We don't optimize specific parameters.

\section{Implementation details for decompositions}\label{apx:decompo_details}

We used the SVD implementation in \texttt{sparse.linalg.svds} provided in the \texttt{Scipy} package. This function makes use of an incremental strategy to only estimate the first singular vectors of $\va$, making the truncated SVD scalable to large matrices. Note that this implementation turned out to obtain the most robust decompositions of $\va$ after an empirical comparison between  \texttt{torch.svd}, \texttt{torch.linalg.svd}, \texttt{scipy.linalg.svd}, and \texttt{sklearn.decomposition.TruncatedSVD} on our data. 

We used PCA and ICA implementations in \texttt{decomposition.PCA} and \texttt{decomposition.FastICA} respectivly provided in the \texttt{scikit-learn} package.

\section{Sobol Indices - details}\label{apx:sobol}

In this section, we define the classic Total Sobol indices and how they are calculated using our method. In practice, these indices can be calculated very efficiently~\cite{marrel2009calculations,saltelli2010variance,janon2014asymptotic} with Quasi-Monte Carlo sampling and the estimator explained below.

Let $(\Omega, \mathcal{A}, \mathbb{P})$ be a probability space of possible concept perturbations. To build these concept perturbations, we use $\boldsymbol{M}=\left(M_1, \ldots, M_r\right) \in \mathcal{M} \subseteq[0,1]^r$, i.i.d. stochastic masks where $M_i \sim \mathcal{U}(\left[  0,1 \right]),  \forall i \in 1,...,r$. We define concept perturbation $\Tilde{\vu}=\boldsymbol{\pi}(\vu, \boldsymbol{M})$ with the perturbation operator $\boldsymbol{\pi}(\vu, \boldsymbol{M})=\vu \odot \boldsymbol{M}$ with $\odot$ the Hadamard product (that takes in two matrices of the same dimensions and returns a matrix of the multiplied corresponding elements).

We denote the set $\mathcal{U}=\{1, \ldots, r\}$, $\boldsymbol{u}$ a subset of $\mathcal{U}$, its complementary $\sim \boldsymbol{u}$ and $\mathbb{E}(\cdot)$ the expectation over the perturbation space. We define $\phi: \mathcal{A} \rightarrow \mathbb{R}$, a function that takes the perturbations $\vu$ from the last layer and applies the difference between the model's two largest output logits -- i.e. $\phi(\vu) = l_{(1)}(c(\vu))- l_{(2)}(c(\vu))$ where $c$ is the classification head of the model (when we do Sobol for the occupation task, it is $c$, and when we do Sobol for the gender task, it is $c_g$) and $l_{(1)}$ and $l_{(2)}$ represent the highest and second highest logit values respectively. We assume that $\phi\in \mathbb{L}^2(\mathcal{A}, \mathbb{P})$ -- i.e. $|\mathbb{E}(\phi(\boldsymbol{U}))|<+\infty$.

The Hoeffding decomposition provides $\phi$ as a function of summands of increasing dimension, denoting $\phi_{\boldsymbol{u}}$ the partial contribution of the concepts $\boldsymbol{U}_{\boldsymbol{u}}=\left(U_i\right)_{i \in \boldsymbol{u}}$ to the score $\phi(\boldsymbol{U})$:

\begin{equation}
\label{eq:hoeffding}
\begin{aligned}
\phi(\boldsymbol{U}) & =\phi_{\varnothing} \\
& +\sum_{i=1}^r \phi_i\left(U_i\right) \\
& +\sum_{1 \leqslant i<j \leqslant r} \phi_{i, j}\left(U_i, U_j\right)+\cdots \\
& +\phi_{1, \ldots, r}\left(U_1, \ldots, U_r\right) \\
& =\sum_{\boldsymbol{u} \subseteq \mathcal{U}} \phi_{\boldsymbol{u}}\left(\boldsymbol{U}_{\boldsymbol{u}}\right)
\end{aligned}
\end{equation}

Eq.~\ref{eq:hoeffding} consists of $2^r$ terms and is unique under the orthogonality constraint:
$$
\quad \mathbb{E}\left(\phi_{\boldsymbol{u}}\left(\boldsymbol{U}_{\boldsymbol{u}}\right) \phi_{\boldsymbol{v}}\left(\boldsymbol{U}_{\boldsymbol{v}}\right)\right)=0,   
$$
$$ \forall(\boldsymbol{u},\boldsymbol{v}) \subseteq \mathcal{U}^2\text { s.t. } \boldsymbol{u} \neq \boldsymbol{v}$$
Moreover, thanks to orthogonality, we have $\phi_{\boldsymbol{u}}\left(\boldsymbol{U}_{\boldsymbol{u}}\right)=\mathbb{E}\left(\phi(\boldsymbol{U}) \mid \boldsymbol{U}_{\boldsymbol{u}}\right)-\sum_{\boldsymbol{v} \subset \boldsymbol{u}} \phi_{\boldsymbol{v}}\left(\boldsymbol{U}_{\boldsymbol{v}}\right)$ and we can write the model variance as:

\begin{equation}
\label{eq:var}
    \begin{aligned}
\mathbb{V}(\phi(\boldsymbol{U})) & =\sum_i^r \mathbb{V}\left(\phi_i\left(U_i\right)\right) \\
& +\sum_{1 \leqslant i<j \leqslant r} \mathbb{V}\left(\phi_{i, j}\left(U_i, U_j\right)\right) \\
& +\ldots+\mathbb{V}\left(\phi_{1, \ldots, r}\left(U_1, \ldots, U_r\right)\right) \\
& =\sum_{\boldsymbol{u} \subseteq \mathcal{U}} \mathbb{V}\left(\phi_{\boldsymbol{u}}\left(\boldsymbol{U}_{\boldsymbol{u}}\right)\right)
\end{aligned}
\end{equation}

Eq.~\ref{eq:var} allows us to write the influence of any subset of concepts $\boldsymbol{u}$ as its own variance. This yields, after normalization by $\mathbb{V}(\phi(\boldsymbol{U}))$, the general definition of Sobol' indices.

\paragraph{Definition}\textit{Sobol' indices \cite{sobol1993sensitivity}.}
The sensitivity index $\mathcal{S}_{\boldsymbol{u}}$ which measures the contribution of the concept set $\boldsymbol{U}_{\boldsymbol{u}}$ to the model response $f(\boldsymbol{U})$ in terms of fluctuation is given by:
\begin{equation}
\label{eq:sobol}
\mathcal{S}_{\boldsymbol{u}} =\frac{\mathbb{V}\left(\phi_{\boldsymbol{u}}\left(\boldsymbol{U}_{\boldsymbol{u}}\right)\right)}{\mathbb{V}(\phi(\boldsymbol{U}))}= \\
\end{equation}
$$\frac{\mathbb{V}\left(\mathbb{E}\left(\phi(\boldsymbol{U}) \mid \boldsymbol{U}_{\boldsymbol{u}}\right)\right)-\sum_{\boldsymbol{v} \subset \boldsymbol{u}} \mathbb{V}\left(\mathbb{E}\left(\phi(\boldsymbol{U}) \mid \boldsymbol{U}_{\boldsymbol{v}}\right)\right)}{\mathbb{V}(\phi(\boldsymbol{U}))}$$

Sobol' indices provide a numerical assessment of the importance of various subsets of concepts in relation to the model's decision-making process. Thus, we have: $\sum_{\boldsymbol{u} \subseteq \mathcal{U}} \mathcal{S}_{\boldsymbol{u}}=1$.

Additionally, the use of Sobol' indices allows for the efficient identification of higher-order interactions between features. Thus, we can define the Total Sobol indices as the sum of of all the Sobol indices containing the concept $i: \mathcal{S}_{T_i}=\sum_{\boldsymbol{u} \subseteq \mathcal{U}, i \in \boldsymbol{u}} \mathcal{S}_{\boldsymbol{u}}$. 
So we can write:
\paragraph{Definition}\textit{Total Sobol indices.}
The total Sobol index $\mathcal{S}{T_i}$, which measures the contribution of a concept $\vu_i$ as well as its interactions of any order with any other concepts to the model output variance, is given by:
\begin{align}
        \label{eq:total_sobol}
           \mathcal{S}_{T_i} 
           & = \frac{ \mathbb{E}_{\bm{M}_{\sim i}}( \Var_{M_i} ( \vy | \bm{M}_{\sim i} )) }{ \Var(\vy) } \\
           & = \frac{ \mathbb{E}_{\bm{M}_{\sim i}}( \Var_{M_i} ( \phi((\vu \odot \vm)\vw) | \bm{M}_{\sim i} )) }{ \Var( \phi((\vu \odot \vm)\vw)) }.
    \end{align}

In practice, our implementation of this method remains close to what is done in COCKATIEL \cite{jourdan2023cockatiel}, with the exception of changes to the $\phi$ function created here to manage concepts found in all the classes at the same time (unlike COCKATIEL, which concentrated on finding concepts in a per-class basis).

\begin{figure*}[p]
    \centering
    \begin{subfigure}{.5\textwidth}
        \centering
        \includegraphics[width=1\linewidth]{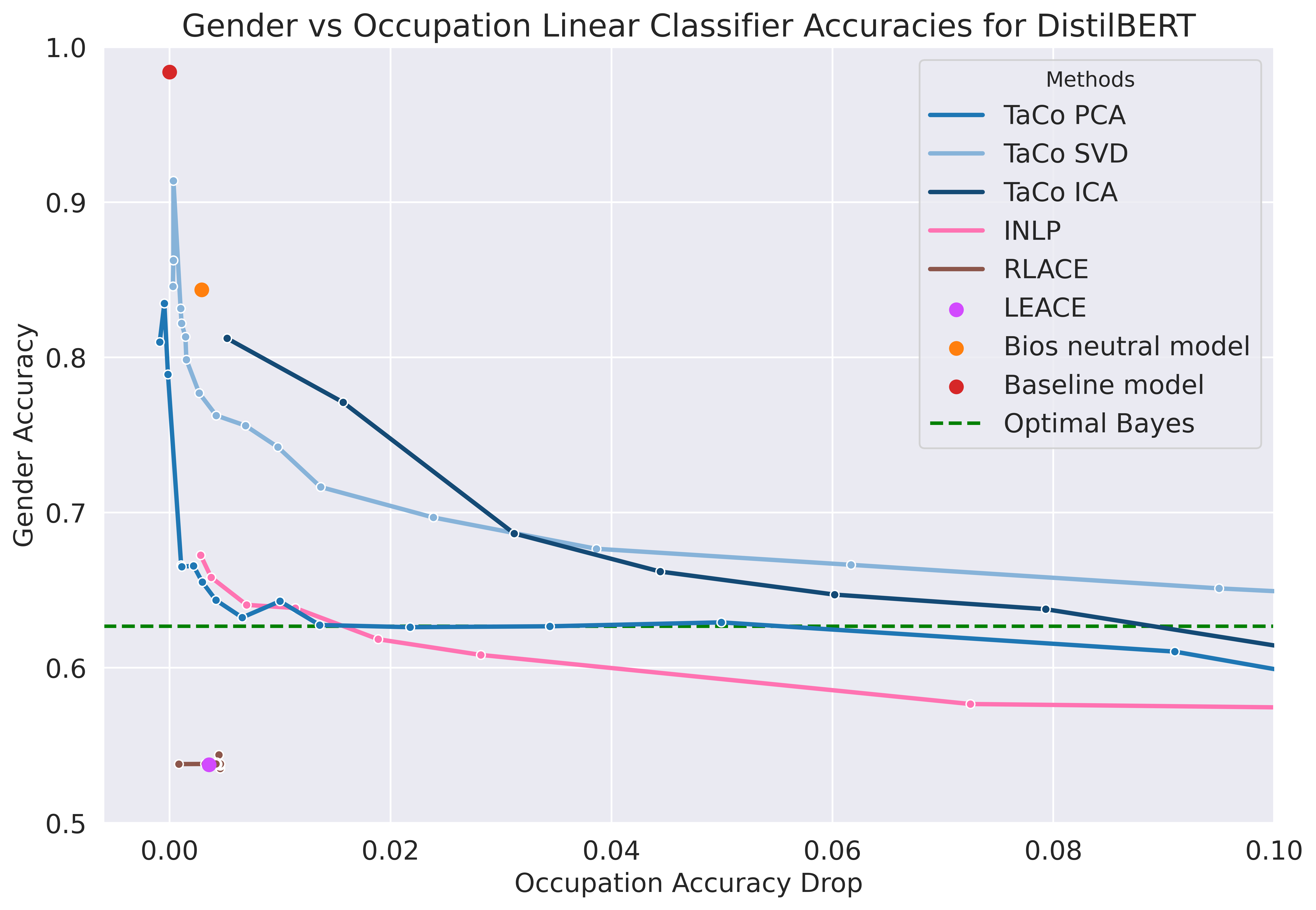}
    \end{subfigure}%
    \hspace*{\fill}
    \begin{subfigure}{.5\textwidth}
        \centering
        \includegraphics[width=1\linewidth]{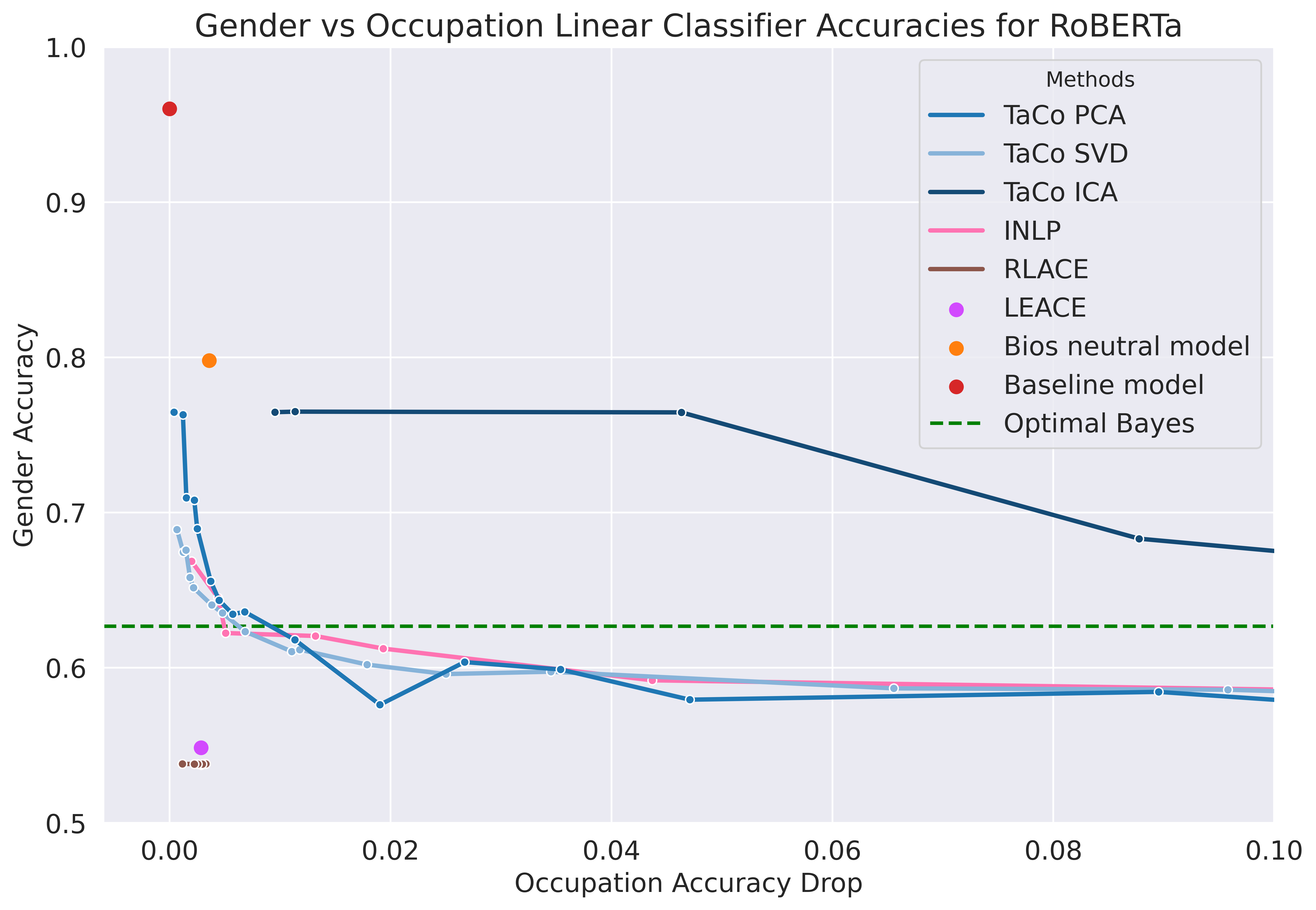}
    \end{subfigure}
    \vspace{0.5cm} 
    \begin{subfigure}{.5\textwidth}
        \centering
        \includegraphics[width=1\linewidth]{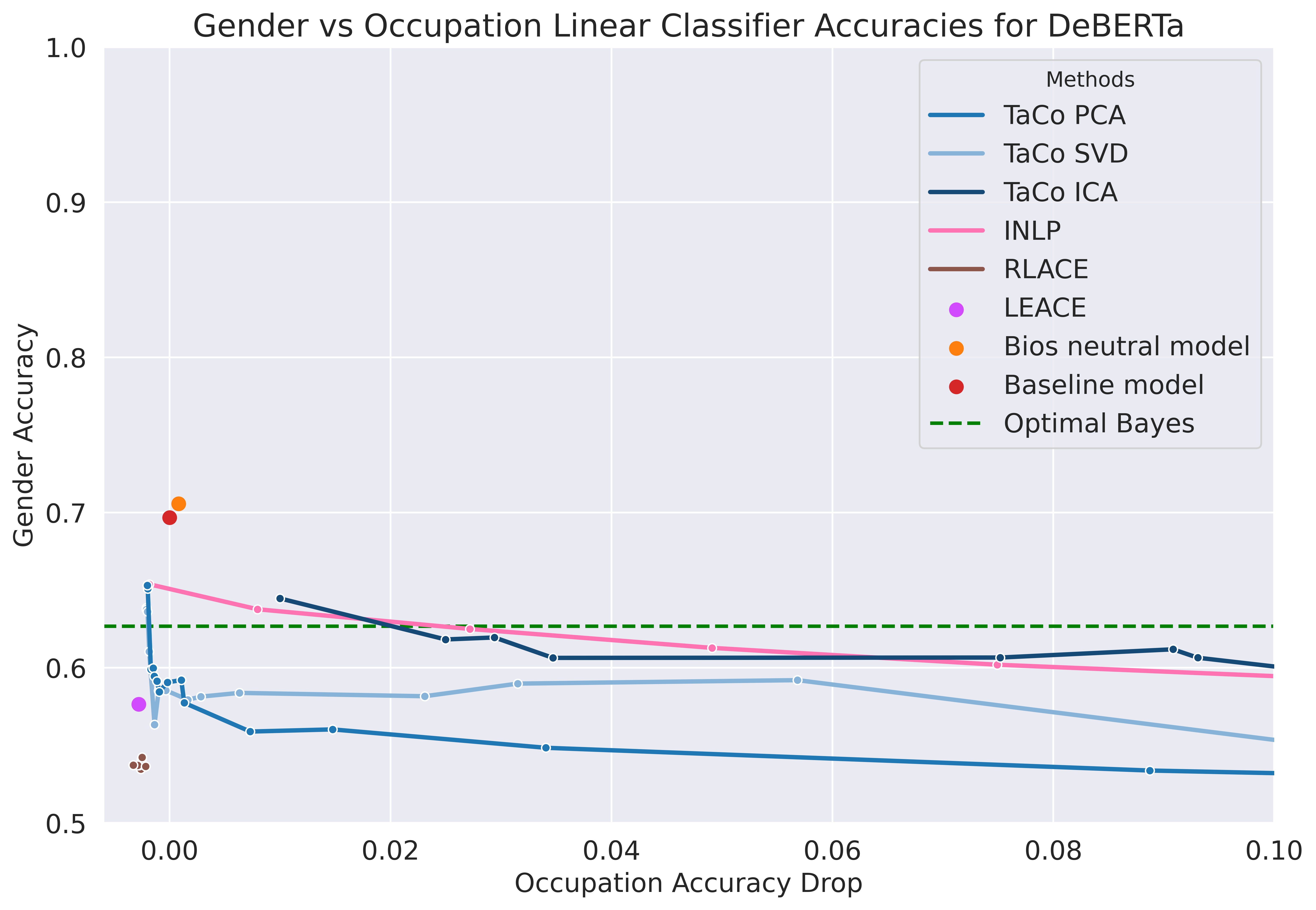}
    \end{subfigure}%
    \hspace*{\fill}
    \begin{subfigure}{.5\textwidth}
        \centering
        \includegraphics[width=1\linewidth]{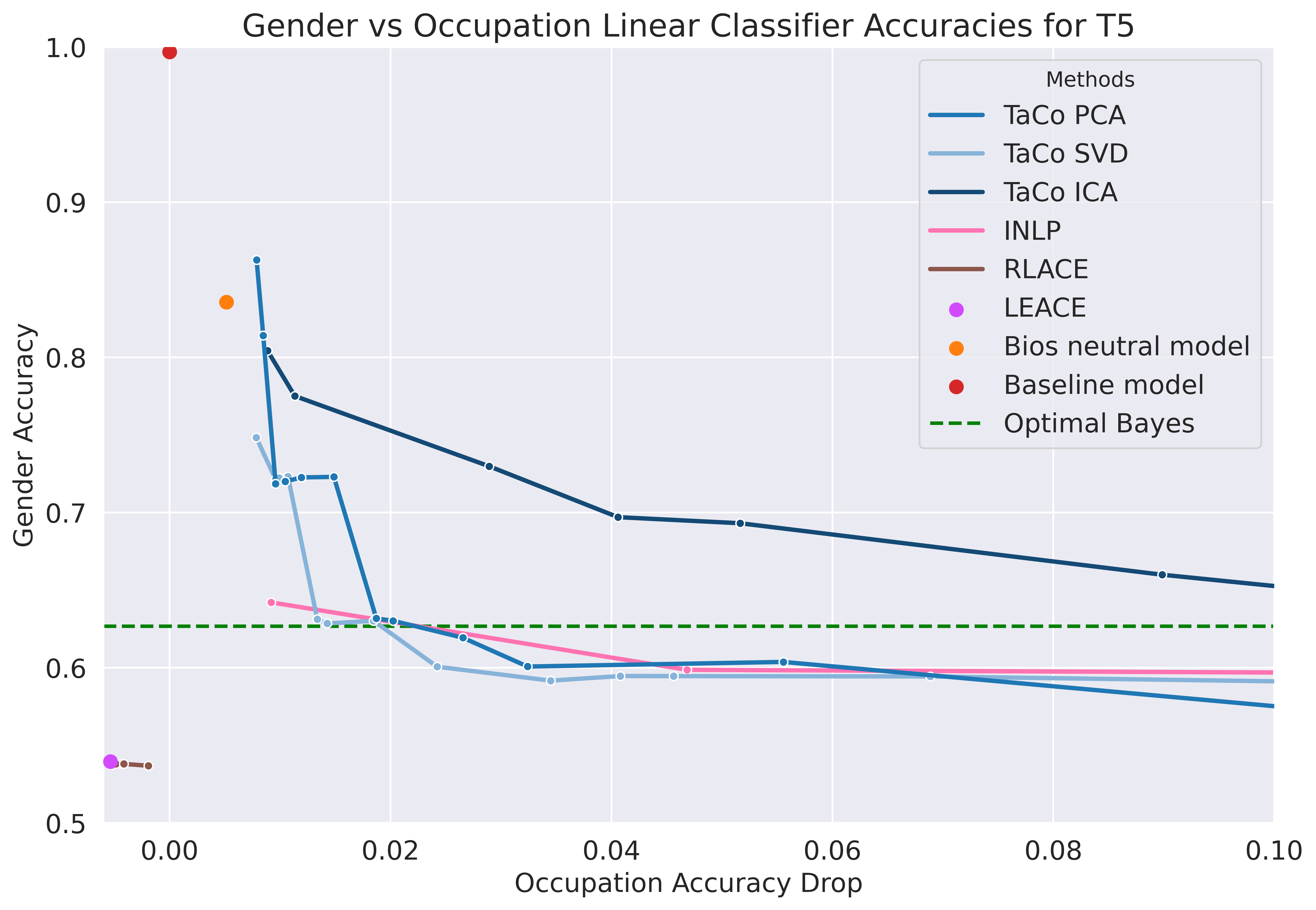}
    \end{subfigure}

    \caption{\textbf{Gender prediction accuracy versus the occupation accuracy drop -- using a Logistic Regression -- after concept erasure methods on (top left) DistilBERT, (top right) RoBERTa, (bottom left) DeBERTa, and (bottom right) T5 representations.}
    The occupation accuracy drop is reported relatively to the \texttt{Baseline model}, when the value is negative for a method, this means that occupation accuracy is better with the method than it was initially.
    For INLP and RLACE, points represent different numbers of masked dimensions.
    For TaCo methods, points represent different numbers of removed concepts. LEACE and the \textit{Bios-neutral} model are shown as single points, as they do not require parameter adjustments. The horizontal dashed line represents the accuracy of the Optimal Bayes classifier, indicating the theoretical lower bound for gender prediction accuracy.}
    \label{apx:fig:linear_results}
\end{figure*}

\begin{figure*}[p]
    \centering
    \begin{subfigure}{.5\textwidth}
        \centering
        \includegraphics[width=1\linewidth]{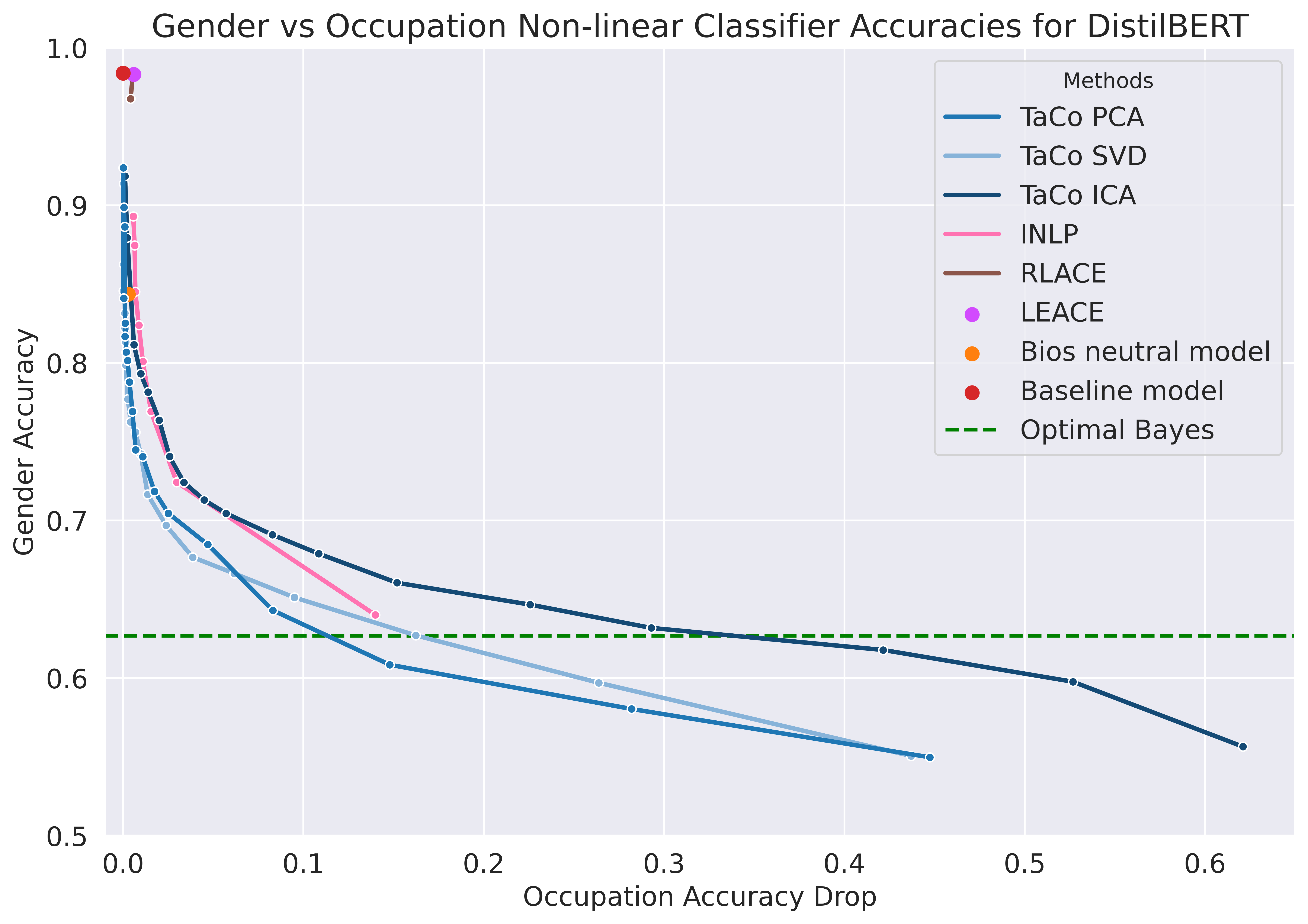}
    \end{subfigure}%
    \hspace*{\fill}
    \begin{subfigure}{.5\textwidth}
        \centering
        \includegraphics[width=1\linewidth]{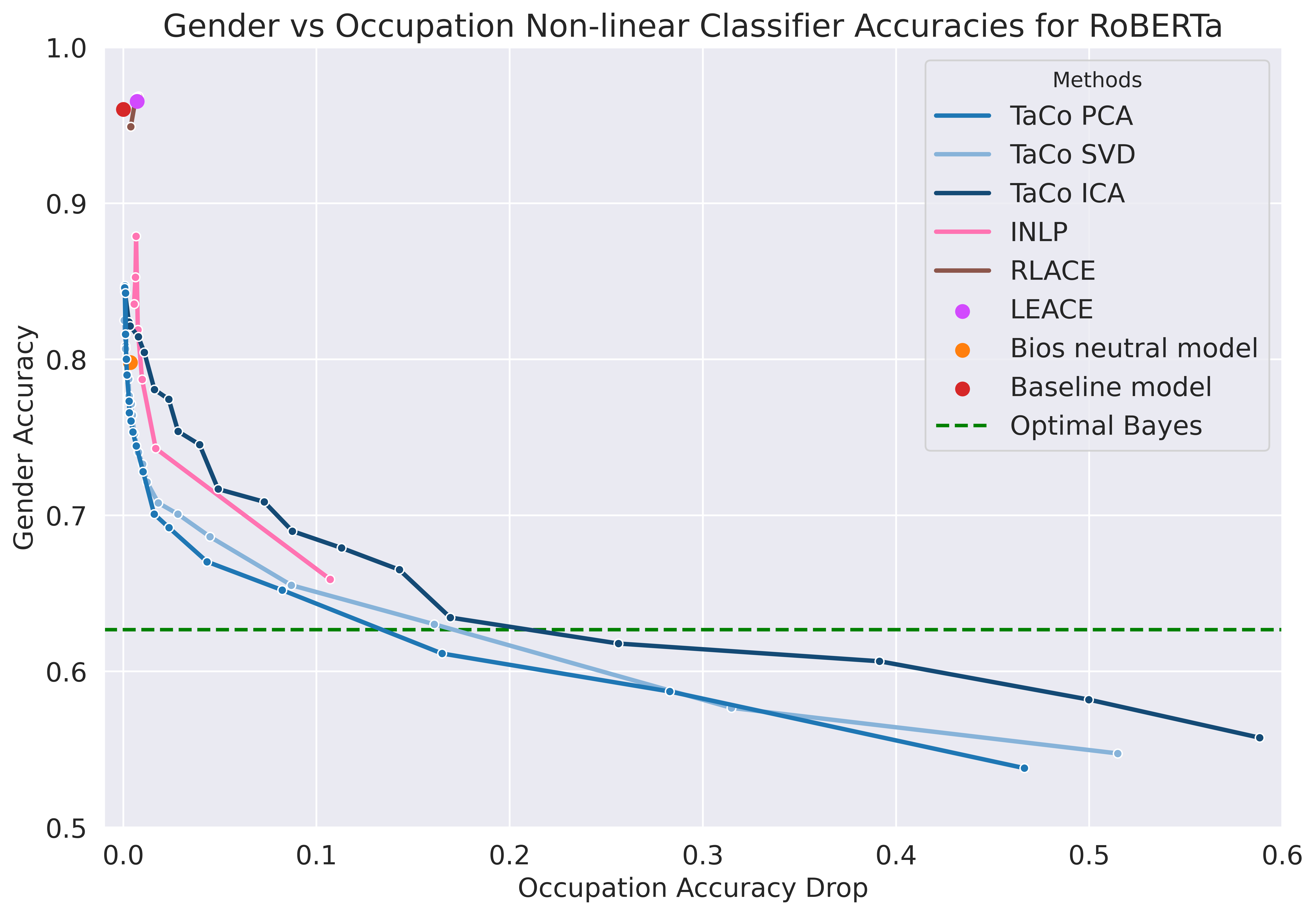}
    \end{subfigure}
    \vspace{0.5cm} 
    \begin{subfigure}{.5\textwidth}
        \centering
        \includegraphics[width=1\linewidth]{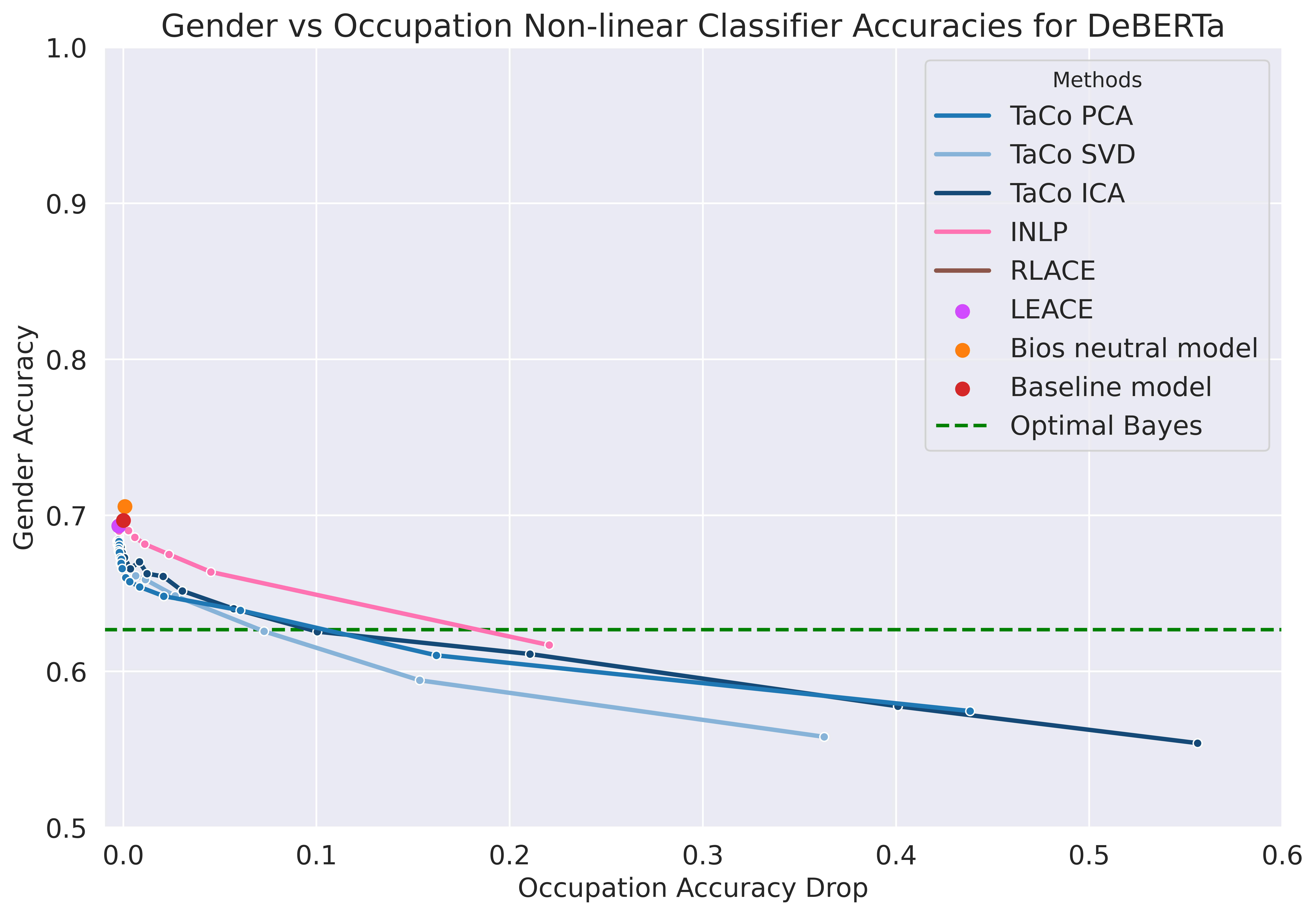}
    \end{subfigure}%
    \hspace*{\fill}
    \begin{subfigure}{.5\textwidth}
        \centering
        \includegraphics[width=1\linewidth]{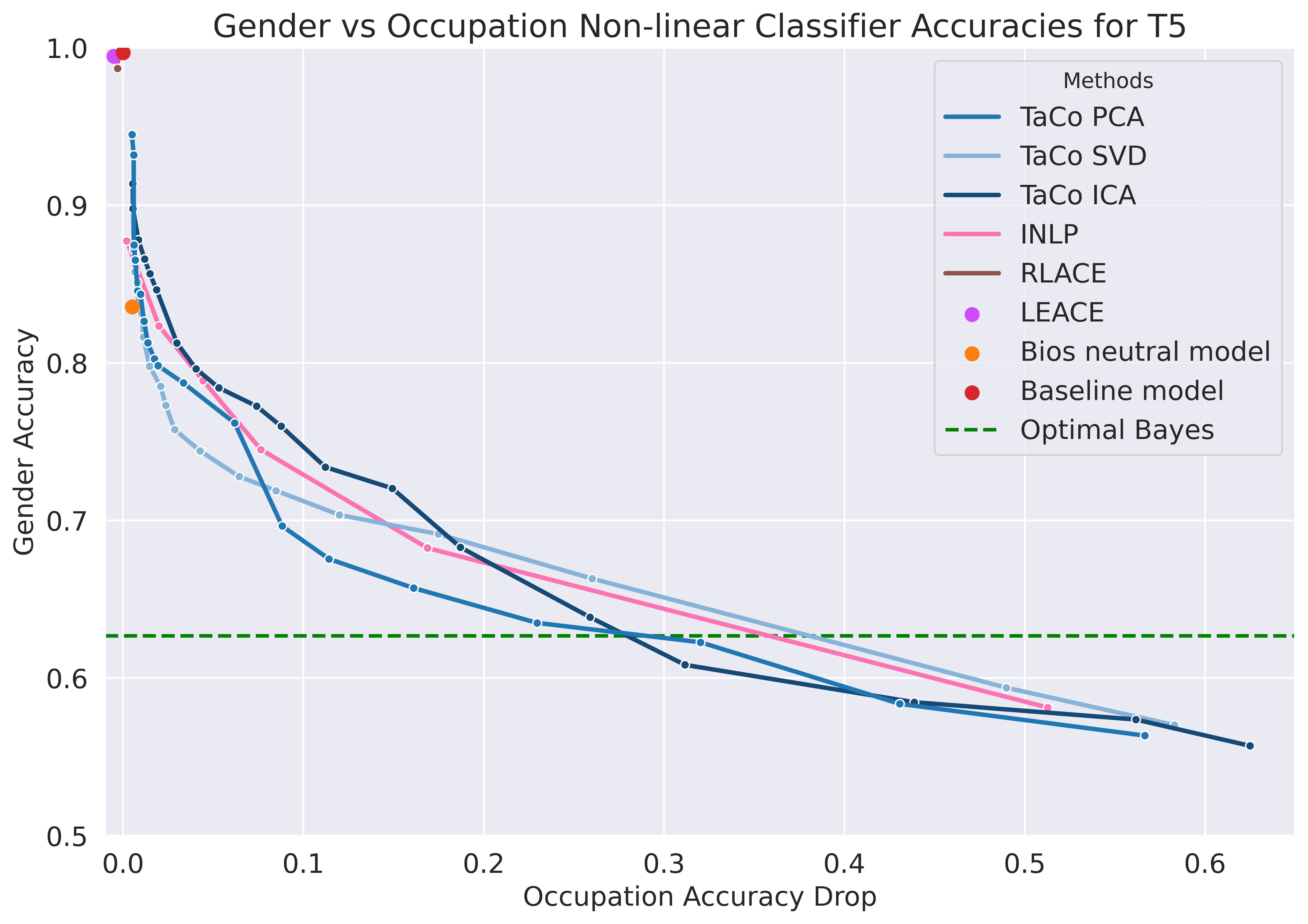}
    \end{subfigure}

    \caption{\textbf{Gender prediction accuracy versus the occupation accuracy drop -- using a two-layer MLP -- after concept erasure methods on (top left) DistilBERT, (top right) RoBERTa, (bottom left) DeBERTa, and (bottom right) T5 representations.}
    The occupation accuracy drop is reported relatively to the \texttt{Baseline model}, when the value is negative for a method, this means that occupation accuracy is better with the method than it was initially.
    For INLP and RLACE, points represent different numbers of masked dimensions.
    For TaCo methods, points represent different numbers of removed concepts. LEACE and the \textit{Bios-neutral} model are shown as single points, as they do not require parameter adjustments. The horizontal dashed line represents the accuracy of the Optimal Bayes classifier, indicating the theoretical lower bound for gender prediction accuracy.}
    \label{apx:fig:nonlinear_results_all}
\end{figure*}

\begin{figure*}[p]
    \centering
    \begin{subfigure}{.5\textwidth}
        \centering
        \includegraphics[width=1\linewidth]{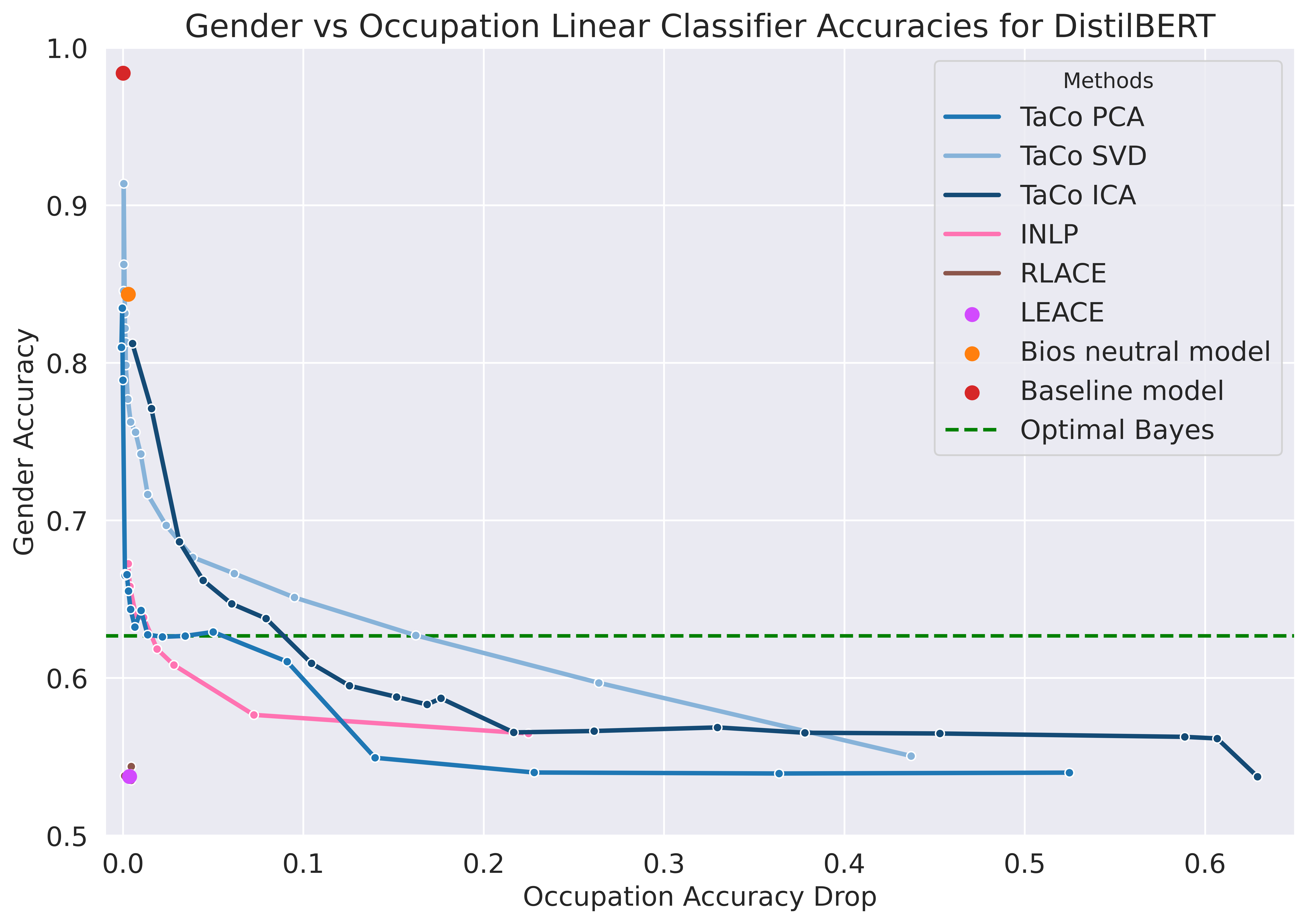}
    \end{subfigure}%
    \hspace*{\fill}
    \begin{subfigure}{.5\textwidth}
        \centering
        \includegraphics[width=1\linewidth]{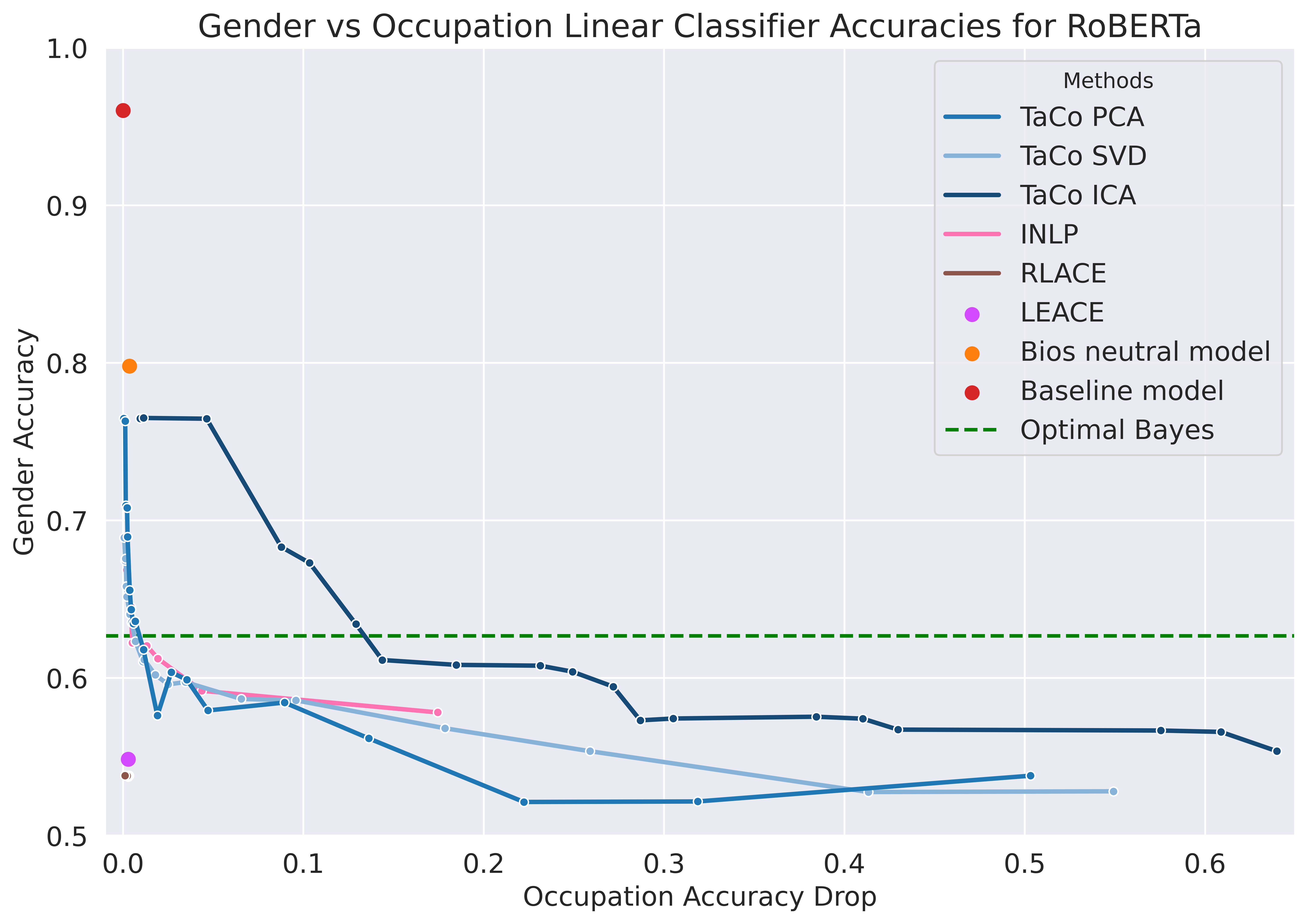}
    \end{subfigure}
    \vspace{0.5cm} 
    \begin{subfigure}{.5\textwidth}
        \centering
        \includegraphics[width=1\linewidth]{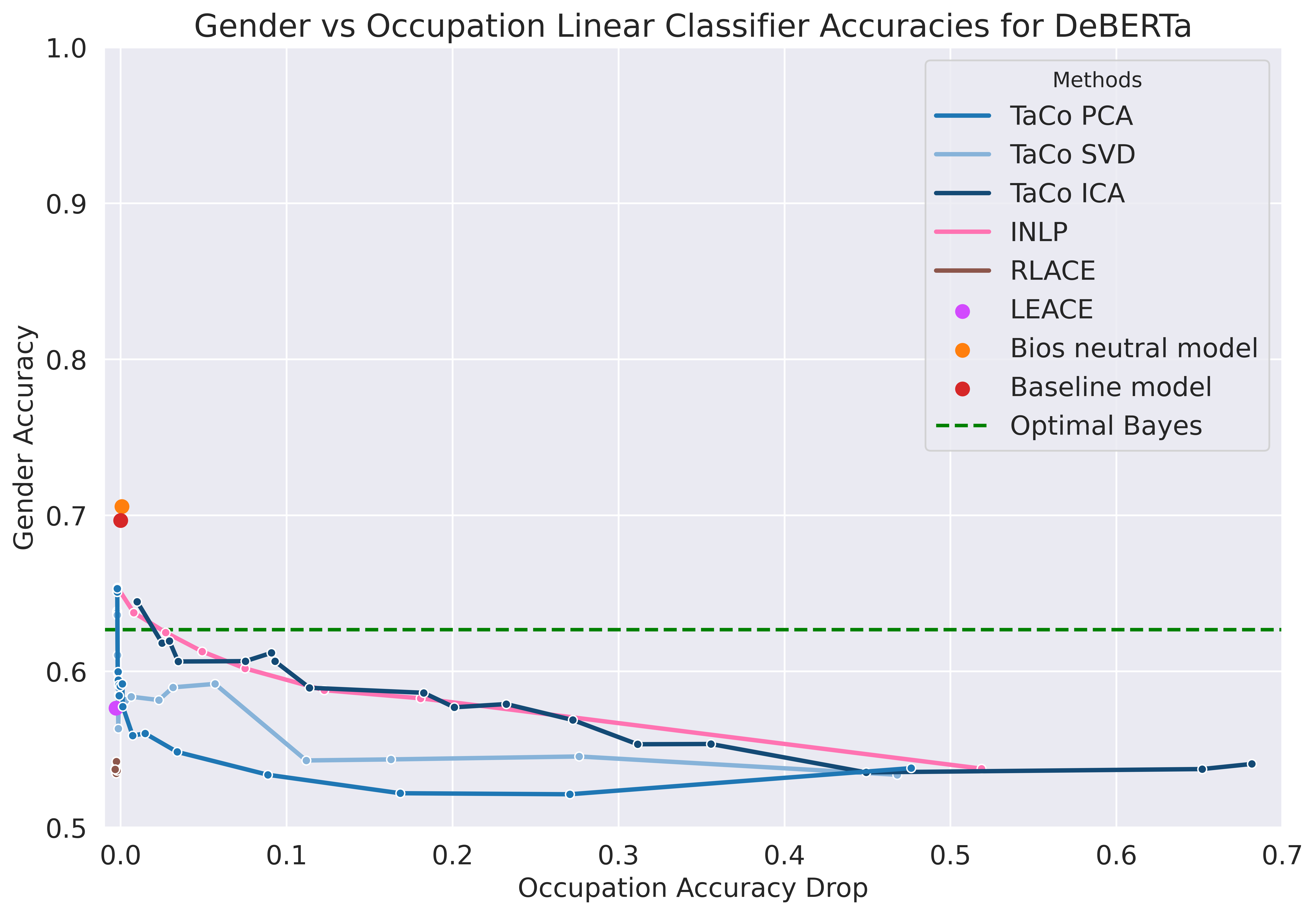}
    \end{subfigure}%
    \hspace*{\fill}
    \begin{subfigure}{.5\textwidth}
        \centering
        \includegraphics[width=1\linewidth]{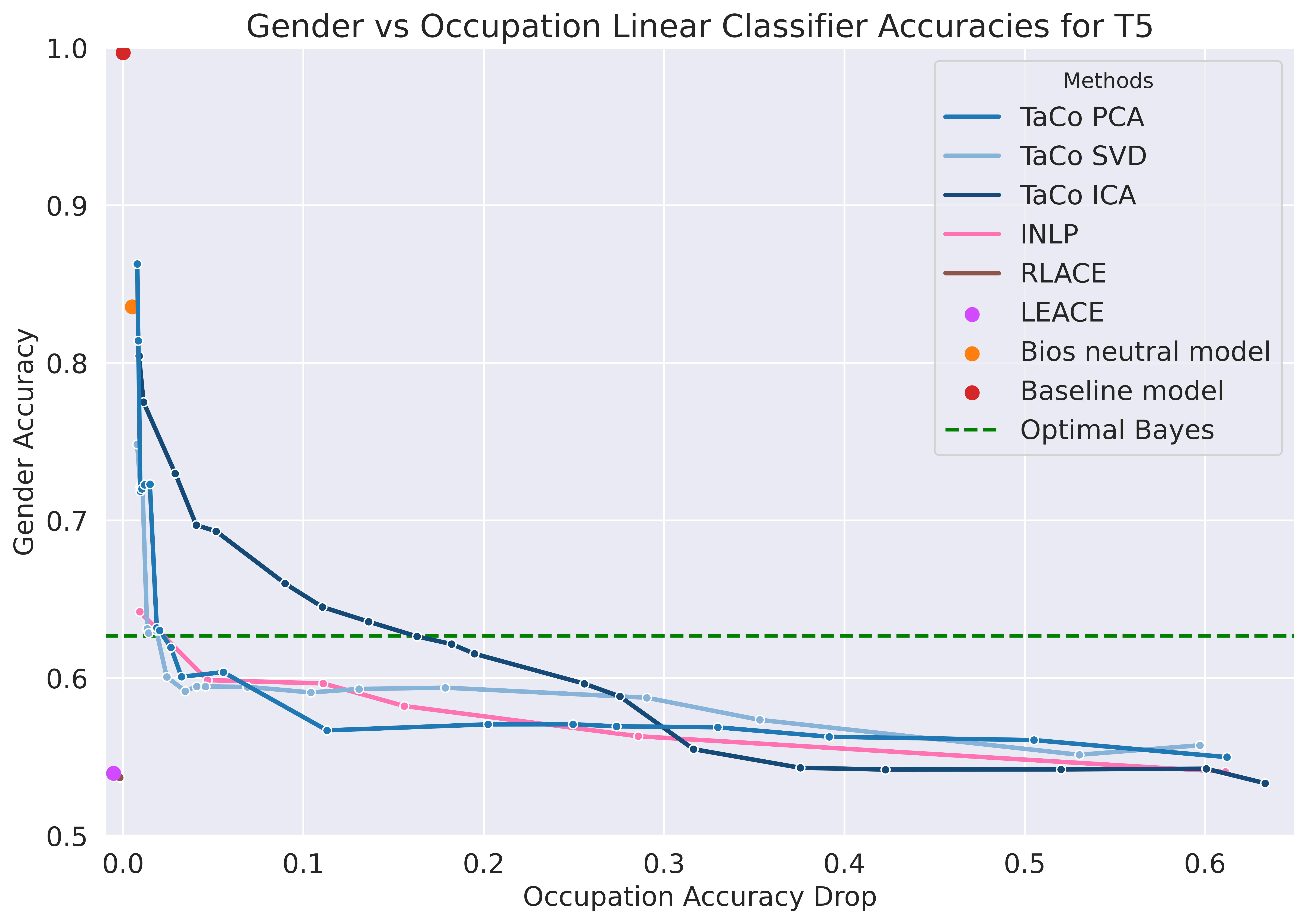}
    \end{subfigure}

    \caption{\textbf{Gender prediction accuracy versus the occupation accuracy drop -- using a Logistic Regression -- after concept erasure methods on (top left) DistilBERT, (top right) RoBERTa, (bottom left) DeBERTa, and (bottom right) T5 representations.}
    The occupation accuracy drop is reported relatively to the \texttt{Baseline model}, when the value is negative for a method, this means that occupation accuracy is better with the method than it was initially.
    For INLP and RLACE, points represent different numbers of masked dimensions.
    For TaCo methods, points represent different numbers of removed concepts. LEACE and the \textit{Bios-neutral} model are shown as single points, as they do not require parameter adjustments. The horizontal dashed line represents the accuracy of the Optimal Bayes classifier, indicating the theoretical lower bound for gender prediction accuracy.}
    \label{apx:fig:linear_results_all}
\end{figure*}

\section{Additional Figures}\label{apx:additionalfig}

Figure \ref{apx:fig:linear_results} presents the results for the linear case, complementing the non-linear results shown in the main paper, thus providing a comprehensive comparison of performance based on the nature of the classifiers used.

Figures \ref{apx:fig:nonlinear_results_all} and \ref{apx:fig:linear_results_all} show the full comparative analysis of the debiasing methods, with the x-axis depicting the drop in occupation prediction accuracy relative to the baseline model and the y-axis representing the gender prediction accuracy.

\newpage

\begin{figure*}
    \centering
\includegraphics[width=1.0\linewidth]{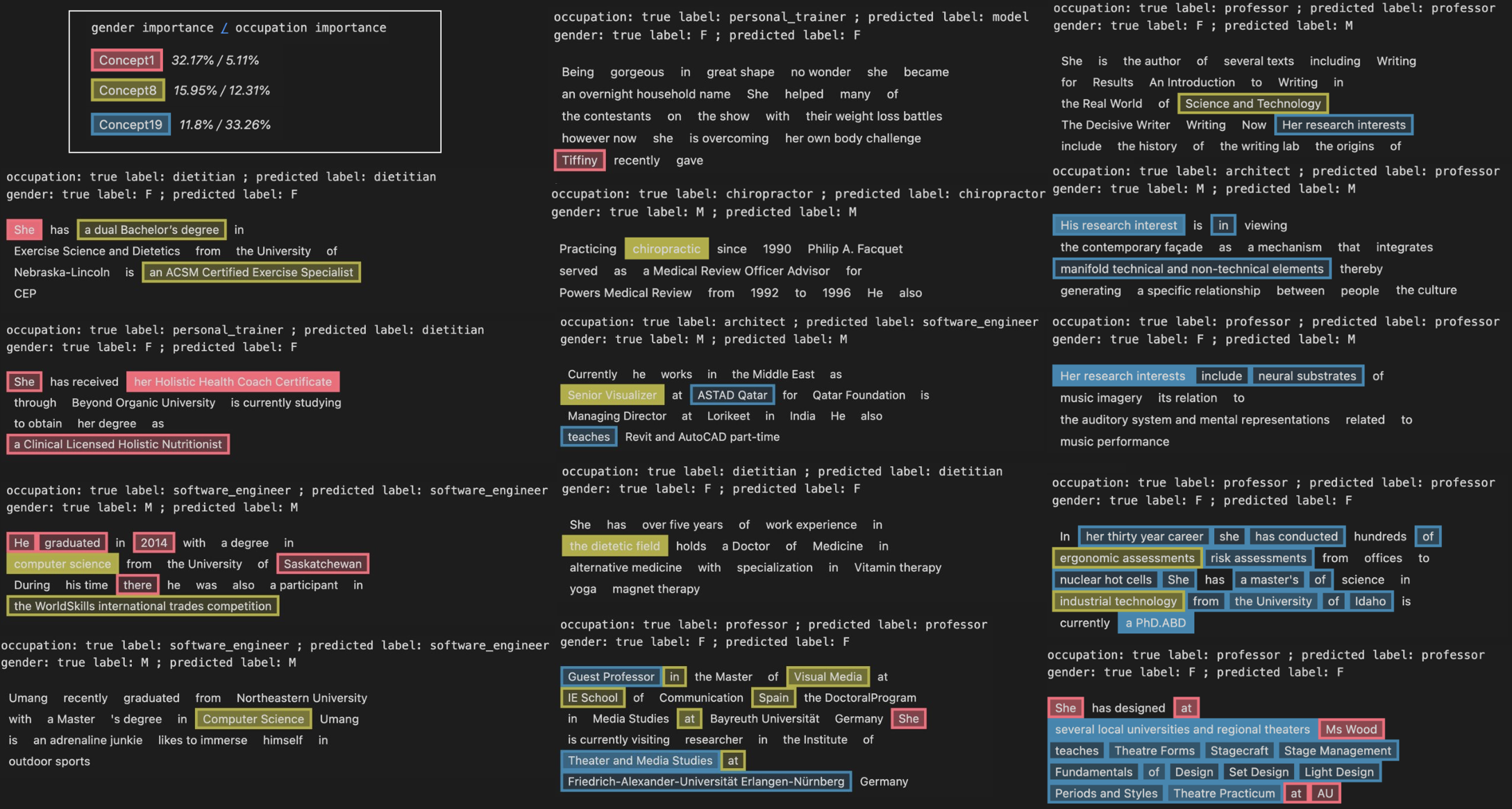}
    \caption{\textbf{Clear interpretability of SVD-Derived Concepts in RoBERTa Using COCKATIEL, showing distinct and meaningful concepts for gender and occupation.} \textbf{\textit{Concept 1}} prominently features explicit gender indicators like '\textit{she}' and '\textit{his}', crucial for gender identification but irrelevant for occupation. \textbf{\textit{Concept 8}} integrates information about gender-imbalanced professions, aiding in predicting both, an individual’s gender and occupation. \textbf{\textit{Concept 19}} focuses on information related to the \textit{professor} occupation, a gender-balanced class and most represented in the dataset.
    \textit{Other explanations for DistilBERT model on Figure \ref{apx:fig:explanationdistilbert}.}}
    \label{apx:fig:explanationroberta}
\end{figure*}

\begin{figure*}
    \centering
\includegraphics[width=1.0\linewidth]{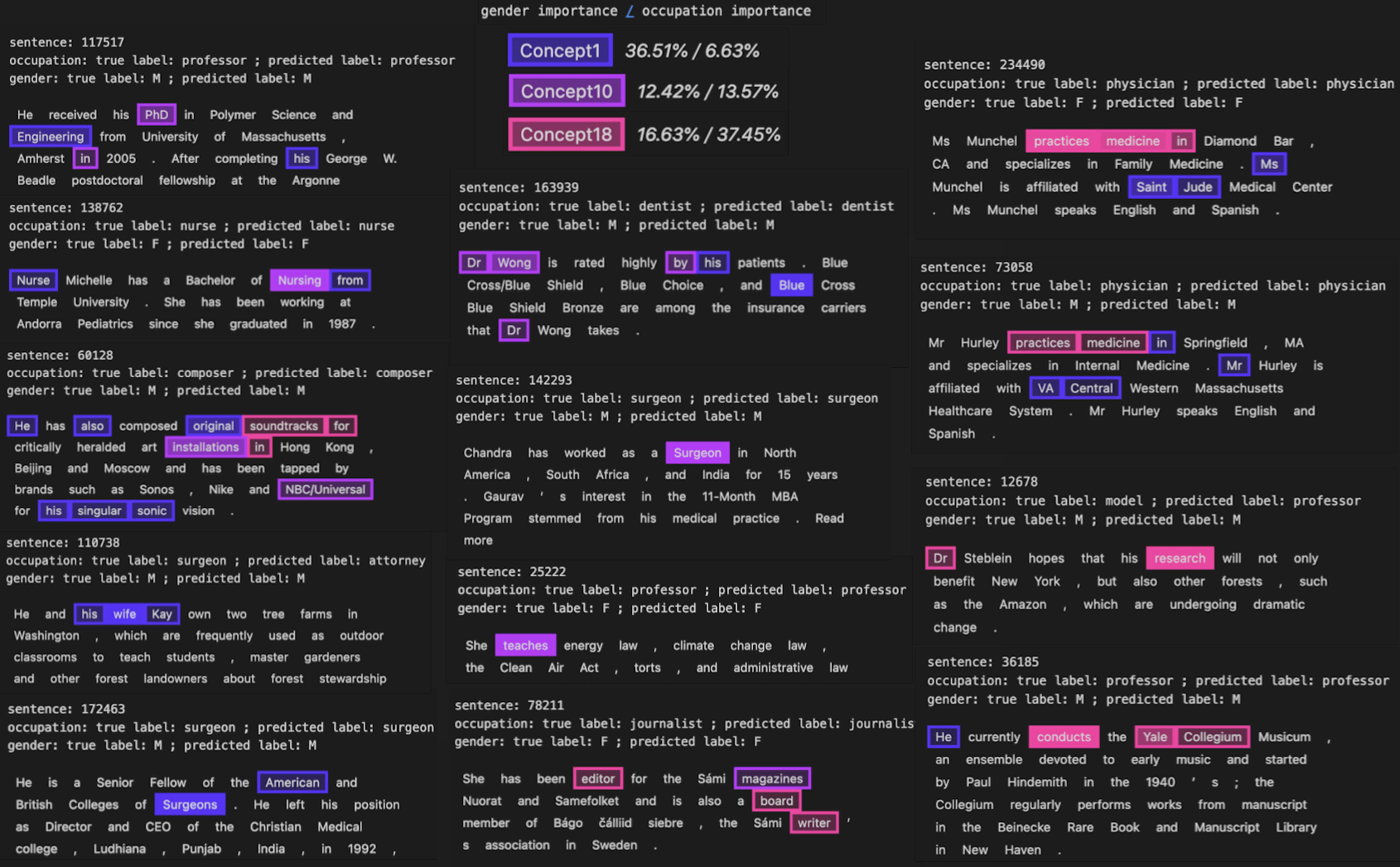}
    \caption{\textbf{Clear interpretability of SVD-Derived Concepts in DistilBERT Using COCKATIEL, with identifiable and interpretable concepts for gender and occupation.}. \textbf{\textit{Concept 1}} prominently features explicit gender indicators like '\textit{she}' and '\textit{his}', crucial for gender identification but irrelevant for occupation, and some information on specific professions that are particularly gender imbalanced. \textbf{\textit{Concept 1O}} integrates information about gender-imbalanced professions, aiding in predicting both, an individual’s gender and occupation, and also some information on occupation in general. \textbf{\textit{Concept 18}} focuses on information related to the occupation in general.}
    \label{apx:fig:explanationdistilbert}
\end{figure*}

\section{Interpretability of Concepts Through Explainability Techniques}\label{apx:explainability}

While the primary focus of our work is on concept erasure for non-linear problems, an additional advantage of our method is its potential to enhance interpretability and transparency in models. This benefit arises from the ability to interpret the concepts identified through various matrix decompositions before they are removed. Our method's construction inherently facilitates this interpretability due to its similarity with explainable AI (XAI) techniques. Such an approach may not be as straightforward with other concept erasure methods.

In this section, we explore this aspect by analyzing the interpretability of concepts derived from Singular Value Decomposition (SVD) and Principal Component Analysis (PCA) across different models representation. Our analysis is facilitated by applying the COCKATIEL method \citep{jourdan2023cockatiel}, which provides explanations for the identified concepts.

\subsection{Common Interpretability Procedure with COCKATIEL}
COCKATIEL modifies the Occlusion method \citep{zeiler2014visualizing} by masking each word in a sentence and observing the resultant impact on the model's output. This approach allows us to infer the significance of each word with respect to a specific concept. By executing this procedure at either the word or clause level, we obtain explanations of varying granularity, enhancing our understanding of how certain concepts influence model predictions. Importantly, this interpretability procedure is uniformly applied across all decompositions and models, enabling a consistent analysis of concepts derived from different methods and architectures.

Our method's structure, which involves decomposing representations into interpretable concepts and ranking them based on their importance to the sensitive attribute and the target label, aligns closely with practices in XAI. This alignment allows us to seamlessly apply interpretability techniques like COCKATIEL to our concepts. Such ease of explanation may not be readily achievable with other concept erasure methods that do not decompose the representation space in a manner conducive to concept-level interpretation.

\subsection{Interpretability of SVD-Derived Concepts}

Using SVD, we decomposed the latent representations and ranked the resulting concepts based on their importance for predicting the sensitive variable (gender) and the target label (occupation) using the Sobol method. We focused on three key concepts for each model, RoBERTa and DistilBERT, that were representative of different importance levels.

\paragraph{RoBERTa Model Analysis} 
\begin{itemize}
    \item Concept 1: High importance for gender (32\%), low for occupation (5\%).
    \item Concept 8: Moderate importance for both gender (16\%) and occupation (12\%).
    \item Concept 19: Low importance for gender (12\%), high for occupation (33\%).
\end{itemize}

As illustrated in Figure \ref{apx:fig:explanationroberta}, applying COCKATIEL to these concepts in the RoBERTa model revealed that: \begin{itemize}
    \item Concept 1 contained explicit gender indicators such as pronouns "\textit{he}" and "\textit{she}," making it highly predictive of gender.
    \item Concept 8 included information from biographies of gender-imbalanced professions (e.g., dietitian, software engineer), thus linking gender and occupation.
    \item Concept 19 focused on the profession of professors—a gender-balanced and highly represented occupation—contributing primarily to occupation prediction without significant gender bias.
\end{itemize}

\paragraph{DistilBERT Model Analysis} 
\begin{itemize}
    \item Concept 1: High importance for gender (36.51\%), low for occupation (6.63\%).
    \item Concept 10: Moderate importance for both gender (12.42\%) and occupation (13.57\%).
    \item Concept 18: Lower importance for gender (16.63\%), high for occupation (37.45\%).
\end{itemize}

Figure \ref{apx:fig:explanationdistilbert} highlights similar findings with the DistilBERT model, although the distinctions between concepts were less pronounced. Nevertheless, the concepts still contained meaningful information:
\begin{itemize}
    \item Concept 1 had explicit gender indicators and terms related to gender-imbalanced occupations.
    \item Concept 10 was informative for both gender and occupation, containing terms from both gender-balanced and imbalanced professions.
    \item Concept 18 primarily focused on occupation-related terms from frequently occurring professions in the dataset.
\end{itemize}

These findings demonstrate that the concepts derived from SVD are interpretable and provide insights into how the model processes information related to gender and occupation. The ability to extract such explanations is facilitated by our method's structure, which is designed to isolate and rank concepts based on their relevance to the sensitive attribute and the target task. This structural design, mirroring XAI techniques, enables straightforward application of interpretability methods like COCKATIEL.

\begin{figure*}
    \centering
\includegraphics[width=1.0\linewidth]{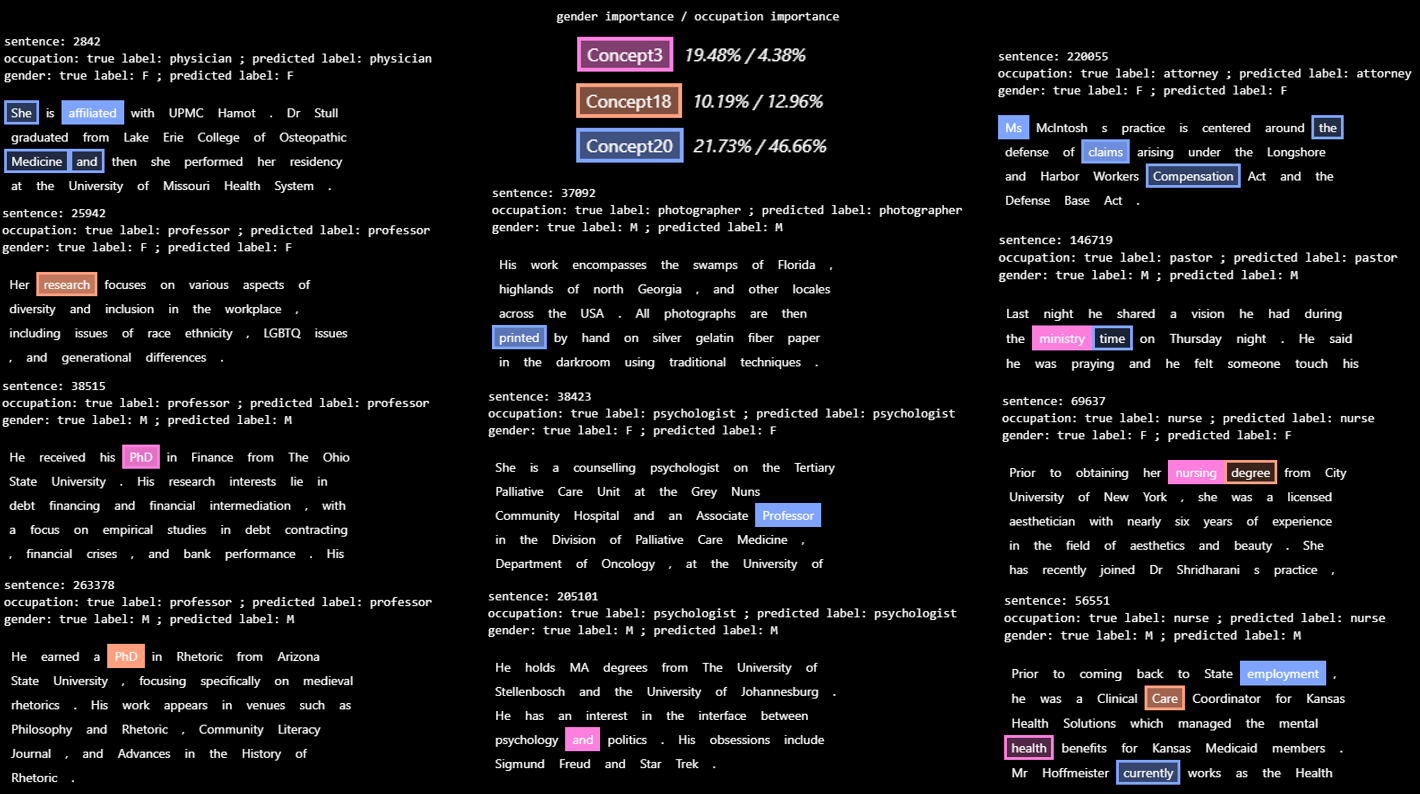}
    \caption{\textbf{Poor interpretability of PCA-Derived Concepts in RoBERTa Using COCKATIEL, revealing less coherent and less meaningful concepts compared to SVD.}}
    \label{apx:fig:explanation_pca}
\end{figure*}

\subsection{Challenges with PCA-Derived Concepts}

The explanations generated by COCKATIEL did not yield coherent or meaningful insights. The concepts lacked clear semantic associations with either gender or occupation, making it difficult to understand their influence on model predictions (see Figure \ref{apx:fig:explanation_pca} for an example on the RoBERTa model).
Despite PCA showing better performance in concept erasure effectiveness, the reduced interpretability of its components presents a challenge. This suggests that while PCA effectively removes sensitive information, the resulting components may not align with human-interpretable concepts. Our method's structure, which emphasizes concept identification and ranking, seems to facilitate interpretability more effectively with decompositions like SVD.

It's important to note that this observation is based on the application of COCKATIEL. Other explainability methods might offer different perspectives and could potentially uncover meaningful interpretations of PCA-derived concepts. However, the ease with which we can apply COCKATIEL to our method, due to its construction that mirrors XAI practices, may not extend to other concept erasure methods that lack this alignment.

\subsection{Implications for Explainability}
The ability to interpret concepts is an additional advantage of our method, providing benefits for model transparency and accountability. The ease with which we can obtain explanations for the concepts is linked to the construction of our method, which is similar to XAI techniques. By decomposing the latent space into ranked concepts, we create a framework that naturally lends itself to interpretability.

This structural alignment allows for straightforward application of interpretability tools like COCKATIEL, which might not be as easily applied to other concept erasure methods that do not decompose representations in a way that isolates individual concepts. As a result, our method not only mitigates the influence of sensitive attributes but also enhances the explainability of the model, offering insights into which concepts are being removed and why.

Our findings suggest that the choice of decomposition and the method's design can impact both the effectiveness of concept erasure and the interpretability of the resulting concepts. The SVD decomposition, combined with our method's structure, facilitates the extraction of meaningful explanations, enhancing transparency.

Given that interpretability is an added advantage rather than the central focus of our work, these findings open avenues for future research. Exploring alternative explainability techniques may enhance the interpretability of PCA-derived concepts or other decompositions. Additionally, investigating other decomposition methods or hybrid approaches could balance the trade-off between concept erasure effectiveness and interpretability.

Our method's structure, which aligns closely with explainability frameworks, allows for such integrations. By facilitating the examination of internal model representations, we contribute to the broader goals of developing AI systems that are not only fairer but also more transparent and interpretable. This added interpretability dimension is a unique strength of our approach and may not be as easily replicated in other concept erasure methods that lack a similar structural design.

\end{document}